\documentclass[journal]{IEEEtran}
\usepackage{graphicx}
\usepackage{amsmath, amssymb}

\usepackage{subcaption}
\usepackage{multirow}
\usepackage{cite}
\usepackage{hyperref}
\usepackage{soul, xcolor}
\usepackage{ulem}
\usepackage{pifont}      
\usepackage{amsmath}     
\usepackage{graphicx}    
\usepackage{booktabs}    
\usepackage{array}
\newcommand{\cmark}{\ding{51}} 
\newcommand{\xmark}{\ding{55}} 
\begin{document}
\title{EnergyFormer: Energy Attention with Fourier Embedding for Hyperspectral Image Classification}
\author{Saad Sohail, Muhammad Usama, Usman Ghous, Manuel Mazzara, Salvatore Distefano, Muhammad Ahmad
\thanks{Saad Sohail, Muhammad Usama, and Usman Ghous are with the Department of Computer Science, National University of Computer and Emerging Sciences, (NUCES), Pakistan. e-mail: (saadsohail5104@gmail.com; m.usama@nu.edu.pk; usman.ghous@nu.edu.pk).}
\thanks{M. Mazzara is with the Institute of Software Development and Engineering, Innopolis University, 420500 Innopolis, Russia (e-mail: m.mazzara@innopolis.ru).}
\thanks{M. Ahmad and S. Distefano are with the Dipartimento di Matematica e Informatica---MIFT, University of Messina, Messina 98121, Italy. (e-mail: mahmad00@gmail.com; sdistefano@unime.it).}
}
\markboth{Journal of \LaTeX\ Class Files}
{Ahmad \MakeLowercase{\textit{et al.}}}
\maketitle
\begin{abstract}
Hyperspectral imaging (HSI) provides rich spectral-spatial information across hundreds of contiguous bands, enabling precise material discrimination in applications such as environmental monitoring, agriculture, and urban analysis. However, the high dimensionality and spectral variability of HSI data pose significant challenges for feature extraction and classification. This paper presents EnergyFormer, a transformer-based framework designed to address these challenges through three key innovations: (1) Multi-Head Energy Attention (MHEA), which optimizes an energy function to selectively enhance critical spectral-spatial features, improving feature discrimination; (2) Fourier Position Embedding (FoPE), which adaptively encodes spectral and spatial dependencies to reinforce long-range interactions; and (3) Enhanced Convolutional Block Attention Module (ECBAM), which selectively amplifies informative wavelength bands and spatial structures, enhancing representation learning. Extensive experiments on the WHU-Hi-HanChuan, Salinas, and Pavia University datasets demonstrate that EnergyFormer achieves exceptional overall accuracies of 99.28\%, 98.63\%, and 98.72\%, respectively, outperforming state-of-the-art CNN, transformer, and Mamba-based models. The source code will be made available at \url{https://github.com/mahmad000}. 
\end{abstract}
\begin{IEEEkeywords}
Hyperspectral Image Classification; Multi-Head Energy Attention; Fourier Position Embedding; Enhanced Convolutional Block Attention.
\end{IEEEkeywords}
\IEEEpeerreviewmaketitle
\section{Introduction}

\IEEEPARstart{H}{yperspectral imaging (HSI)} has transformed numerous fields, including precision agriculture, climate analysis, material discrimination, and environmental monitoring. Its ability to capture fine-grained spectral and spatial details has enabled advancements in target detection, urban planning, and mineral exploration \cite{jiang2025adaptive, xi2025mctgcl}. However, the high dimensionality, spectral variability, and computational complexity of hyperspectral data present significant challenges. Addressing these requires classification models that not only extract spatial-spectral features effectively but also optimize computational efficiency \cite{10102432}. 

To overcome these challenges, vision transformers (ViTs) have emerged as a powerful solution for HSI classification (HSIC), leveraging self-attention mechanisms for precise spectral-spatial modeling \cite{zhu2024center, ahmad2024spatial}. Recent advancements, such as Spectrum Recombine Former (SRF) \cite{jing2025srf} and Spatial-Spectral Transformer (SSFormer) \cite{ahmad2024spatial}, refine feature extraction through cross-layer fusion and conditional position encoding, setting new benchmarks in performance. However, issues like spatial sensitivity, overfitting, and high computational costs remain obstacles. To mitigate these, GSC-ViT \cite{zhao2024hyperspectral} enhances spectral-spatial modeling through parameter optimization, while CCSF-Transformer \cite{zhu2024center} reduces attention redundancy. Furthermore, SClusterFormer \cite{fang2025deformable} advances long-range feature extraction with cluster attention and dual-branch architectures, pushing the boundaries of HSI classification.

Despite these advancements, ViTs still face unresolved challenges that hinder their practical adoption for HSIC. The quadratic complexity of self-attention makes ViTs computationally expensive for high-dimensional HSI data, limiting scalability. Touvron et al. \cite{b16} note that ViTs often lag behind CNNs in data efficiency, especially with scarce labeled samples. Furthermore, spatial redundancy in ViTs, as highlighted by Hu et al. \cite{Hu2024LFViTRS}, inflates memory usage and increases overfitting risks. Fixed patch tokenization and rigid positional encodings further restrict their ability to capture multi-scale spatial-spectral nuances, hindering deployment in resource-constrained devices. While Rotary Positional Embeddings (RoPE) enable Transformers to encode relative positional data, their reliance on a single sinusoidal frequency per dimension limits length generalization \cite{kazemnejad2024impact}. Fixed sinusoidal patterns are prone to spectral damage, exacerbated by linear transformations, activation functions, and time-domain truncation. 

Prior studies have highlighted the limitations of truncation-based positional encoding, where the removal of higher-order frequency components results in undertrained spectral representations, leading to distortions in periodic structures and degradation of long-range dependencies \cite{hua2024fourier}. To address these challenges, Fourier Positional Embedding (FoPE) has been introduced as a more expressive alternative, modeling each positional dimension as a Fourier series. By integrating multiple frequency components, FoPE enhances the richness of positional representations, capturing both local and global dependencies more effectively. Additionally, the incorporation of a floor frequency constraint serves to mitigate spectral artifacts, preserving critical frequency information and improving the model’s ability to generalize across varying sequence lengths.

\begin{figure*}[!hbt]
    \centering
    \includegraphics[width=\linewidth]{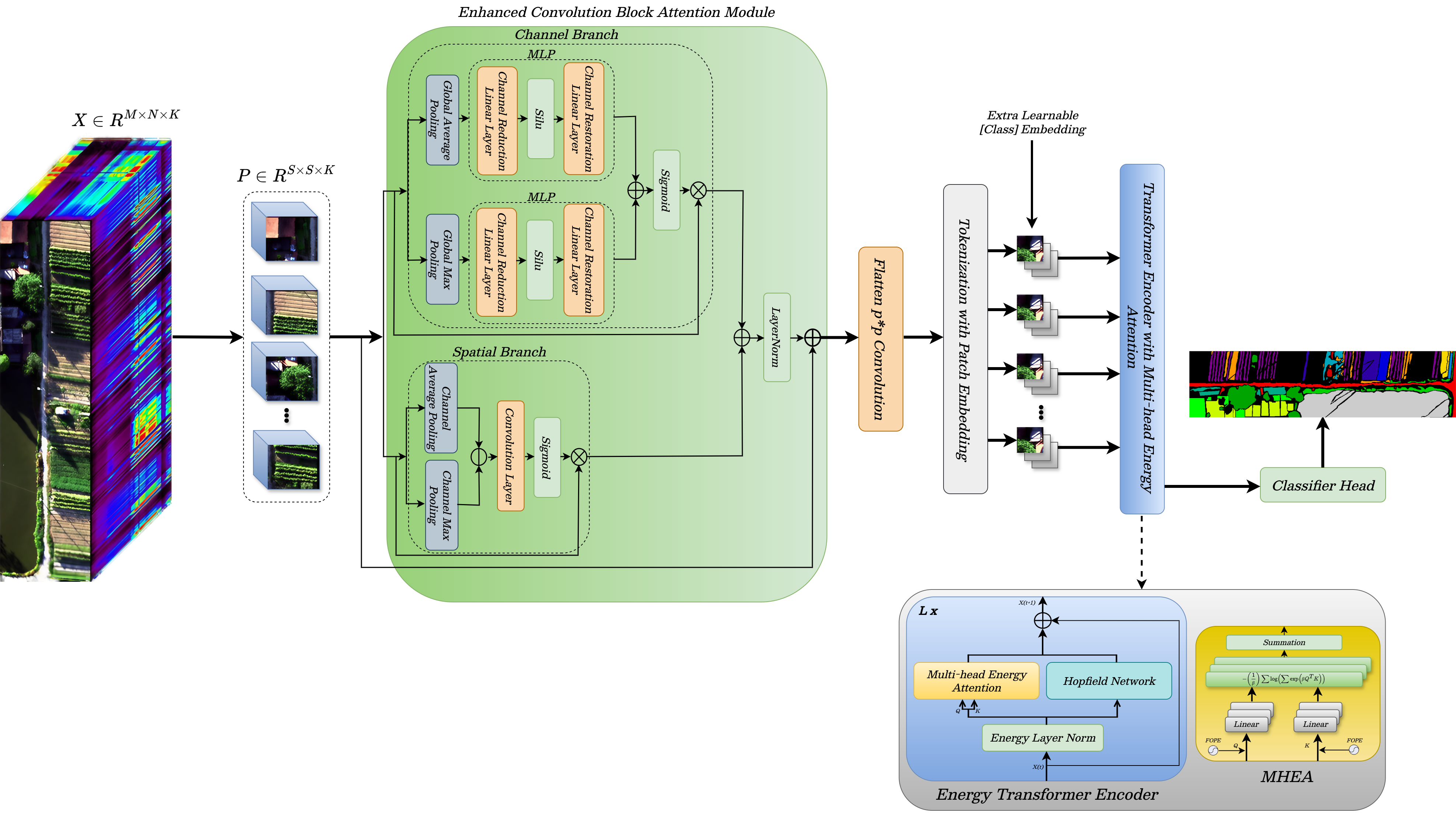}
    \caption{The HSI cube is partitioned into overlapping 3D patches, which are processed by the ECBAM to selectively refine spectral-spatial features. The normalized tokens, incorporating a learnable class token, are then input into the Energy Transformer Encoder, which integrates FoPE for spectral decomposition, MHEA for adaptive feature weighting, and a Hopfield Network for structured feature refinement. The final encoded representations effectively capture both local discriminative details and global contextual dependencies for robust classification.}
\end{figure*}

While FoPE improves positional encoding, effective HSIC demands a more holistic approach that optimally extracts, enhances, and integrates spectral-spatial features. To this end, this work introduces \textit{EnergyFormer}, a novel framework that synergizes Multi-Head Energy Attention (MHEA), FoPE, and an Enhanced Convolutional Block Attention Module (ECBAM) to address the challenges of HSIC comprehensively. MHEA, based on an energy-driven formulation, refines spectral-spatial feature extraction through multiple attention heads, selectively emphasizing critical regions and spectral bands while mitigating noise and atmospheric distortions. Beyond its role in positional encoding, FoPE transforms feature representations by adaptively decomposing spatial and spectral information into multiple frequency components, thereby capturing long-range dependencies and subtle spectral variations. Complementing these, ECBAM further refines feature learning by selectively amplifying discriminative wavelength bands and spatial structures, thereby enhancing class separability. The integration of these three components enables EnergyFormer to mitigate the limitations of state-of-the-art models effectively.

\section{Proposed Methodology}

HSI data consists of $K$ spectral bands with a spatial resolution of $M \times N$, forming an HSI data cube $X \in \mathbb{R}^{M \times N \times K}$. The data cube is partitioned into overlapping 3D patches of size $S \times S \times K$, centered at pixel coordinates $(\alpha, \beta)$. The total number of patches is $m = (M - S + 1) \times (N - S + 1)$. Each patch $P_{\alpha, \beta}$ spans a square region of size $(2L + 1) \times (2L + 1)$, where $L = \lfloor S/2 \rfloor$. Each patch $P_{\alpha, \beta}$ undergoes feature enhancement using an ECBAM, integrating spatial and channel attention mechanisms.  

\textbf{Spatial Attention:} Spatial attention identifies key regions by applying max and average pooling along the channel axis. The concatenated outputs pass through a convolutional layer with a kernel size of $k \times k$, generating an attention map $M_s(X) = \sigma \Big( f_{k \times k} \big( \left[ \text{MaxPool}(X); \text{AvgPool}(X) \right]\big) \Big)$, where $M_s(X) \in \mathbb{R}^{S \times S \times 1}$ and $\sigma$ denotes the sigmoid function. The refined spatial feature representation is:

\begin{equation}
    F_s = M_s(X) \otimes X
\end{equation}
where $\otimes$ denotes element-wise multiplication, broadcast across the channel dimension. \textbf{Channel Attention:} Channel attention captures inter-channel dependencies by applying max and average pooling, followed by an MLP $M_c(X) = \sigma (\text{MLP}(\text{MaxPool}(X)) + \text{MLP}(\text{AvgPool}(X)))$, where the MLP consists of two weight matrices $W_0, W_1$ and a SiLU (Swish) activation function $\delta$ as $M_c(X) = \sigma \big(W_1 \cdot \delta (W_0 \cdot X_{\text{c\_max}} )\big) + W_1 \cdot \delta (W_0 \cdot X_{\text{c\_avg}})$, where $M_c(X) \in \mathbb{R}^{1 \times 1 \times K}$, and the final channel-refined features are:

\begin{equation}
    F_c = M_c(X) \otimes X
\end{equation}

To integrate spatial and channel features effectively, an adaptive fusion is applied $X_{\text{patch}} = \text{LN}\big( X + \gamma \left( \alpha_c \cdot F_c + \alpha_s \cdot F_s \right) \big)$, where $\gamma$ is a learnable scaling factor, $\alpha_c$ and $\alpha_s$ are adaptive weights, and $\text{LN}$ denotes layer normalization. Finally, $X_{\text{patch}}$ is projected into a $d_{\text{embed}}$-dimensional space via a convolutional layer, yielding a feature sequence $X_{\text{patch}} \in \mathbb{R}^{N \times d_{\text{embed}}}$, where $N = S^2$. A trainable class token extends this sequence to $X_{\text{patch}} \in \mathbb{R}^{(N+1) \times d_{\text{embed}}}$.

\paragraph{\textbf{Fourier Positional Encoding (FoPE)}}

To enhance spatial awareness, FoPE modifies the query $(q)$ and key $(k)$ vectors using dominant frequencies $\omega_m$ and learned harmonic coefficients $a_{\omega}$, formulated as:

\begin{equation}
    h_m(n) = H_m(n) \cdot f(\omega_m)
\end{equation}
where the frequency modulation function is:

\begin{equation}
    f(\omega_m) = \begin{cases}
        1, & \text{if } \omega_m < \omega_l \\
        e^{i\omega_m n} + \sum_{\omega} a_{\omega} e^{i\omega n}, & \text{if } \omega_m \geq \omega_l
\end{cases}
\end{equation}
with a lower bound frequency $\omega_l = \frac{2\pi}{N}$ ensuring stability and mitigating spectral distortion.

\paragraph{\textbf{Multi-Head Energy Attention (MHEA)}}

After integrating positional information, MHEA projects the tokens using:

\begin{equation}
    Q^{(h)} = X \cdot W_Q^{(h)}, \quad K^{(h)} = X \cdot W_K^{(h)}
\end{equation}
where $W_Q^{(h)}, W_K^{(h)} \in \mathbb{R}^{d_{\text{model}} \times d}$. Instead of softmax, an energy-based aggregation mechanism is applied:

\begin{equation}
    A_{BC}^{(h)} = \sum_{i=1}^{d} K_{B,i}^{(h)} \cdot Q_{C,i}^{(h)}
\end{equation}

\begin{equation}
    E_{\text{ATT}}^{(h)} = -\frac{1}{\beta} \sum_{C=1}^{N} \log \left( \sum_{\substack{B=1 \\ B \neq C}}^{N} \exp (\beta A_{BC}^{(h)}) \right)
\end{equation}

The total attention energy is:

\begin{equation}
    E_{\text{ATT}} = \sum_{h=1}^{H} E_{\text{ATT}}^{(h)}
\end{equation}

\paragraph{\textbf{Hopfield Network for Token Refinement}}

To refine token representations, a Hopfield Network (HN) replaces the standard feed-forward network. The energy function for token $x$ is defined as:

\begin{equation}
    E_h(x) = -\frac{1}{2} \| \text{ReLU}(W_h x) \|^2
\end{equation}
where $W_h$ is the Hopfield projection matrix. The total block energy is:

\begin{equation}
    E_{\text{total}} = E_{\text{ATT}} + E_h(x)
\end{equation}

Token updates are performed via gradient-based energy minimization:

\begin{equation}
    x^{t+1} = x^t - \alpha \nabla E_{\text{total}}(x^t)
\end{equation}
where $\alpha$ is the learning rate. Energy-based normalization stabilizes updates:

\begin{equation}
    \text{LN}(x) = \gamma \frac{x - \mu}{\sqrt{\sigma^2 + \epsilon}} + \beta
\end{equation}

Finally, the encoder output is processed through a classification head, extracting the class token $z_{\text{cls}}$ to compute logits $y = \text{softmax}(W z_{\text{cls}} + b)$, where $W$ and $b$ are trainable parameters.

\section{Experimental Results and Discussion}

This section presents the experimental results and analysis of EnergyFormer, evaluating its performance on HSIC. We compare its effectiveness against state-of-the-art methods and provide insights into its advantages through quantitative and qualitative assessments. Performance was evaluated using overall accuracy (OA), average accuracy (AA), per-class accuracy, and the $\kappa$ coefficient.

\paragraph{\textbf{Ablation Study}} 

To assess the contribution of each component, we conduct an ablation study by systematically removing one module at a time and evaluating the model's performance. This analysis helps isolate the impact of key design choices, including the MHEA, FoPE, ECBAM, HN, and ELN.

\begin{table}[!htb]
    \centering
    \caption{The contributions of key components, including MHEA, FoPE, ECBAM, HN, and ELN, are analyzed.}
    \label{ablation}
    \resizebox{\linewidth}{!}{\begin{tabular}{lcccccccccc} \hline 
        \multicolumn{10}{c}{\textbf{Dataset 1: Salinas}} \\ \hline 
        \textbf{Model} & \textbf{MHEA} & \textbf{FoPE} & \textbf{ECBAM} & \textbf{HN} & \textbf{ELN} & \textbf{OA (\%)} & \textbf{AA (\%)} & \boldmath$\kappa$ \textbf{(\%)} & \textbf{Time (s)} \\ \hline 
        Full Model & \cmark & \cmark & \cmark & \cmark & \cmark & 98.73 & 99.09 & 98.58 & 436.33 \\
        w/ Std. Attn. + FF & \xmark & \cmark & \cmark & \xmark & \cmark & 98.28 & 98.57 & 98.09 & 389.64 \\
        w/ Std. Attn. + FF + LN & \xmark & \cmark & \cmark & \xmark & \xmark & 97.25 & 97.89 & 96.94 & 566.77 \\ \hline 
        
        \multicolumn{10}{c}{\textbf{Dataset 2: Pavia University}} \\
        \midrule
        \textbf{Model} & \textbf{MHEA} & \textbf{FoPE} & \textbf{ECBAM} & \textbf{HN} & \textbf{ELN} & \textbf{OA (\%)} & \textbf{AA (\%)} & \boldmath$\kappa$ \textbf{(\%)} & \textbf{Time (s)} \\
        \midrule
        Full Model         & \cmark & \cmark & \cmark & \cmark & \cmark & 98.72 & 97.98 & 98.31 & 347.71 \\
        w/ Standard Attention + Feed Forward & \xmark & \cmark & \cmark & \xmark & \cmark & 98.28 & 97.01 & 97.71 & 512.84 \\
        w/ Standard Attention + Feed Forward + LN & \xmark & \cmark & \cmark & \xmark & \xmark & 96.33 & 93.52 & 95.13  & 429.12 \\ \hline 
        \multicolumn{10}{c}{\textbf{Dataset 3: WHU-Hi-HanChuan}} \\ \hline 
        \textbf{Model} & \textbf{MHEA} & \textbf{FoPE} & \textbf{ECBAM} & \textbf{HN} & \textbf{ELN} & \textbf{OA (\%)} & \textbf{AA (\%)} & \boldmath$\kappa$ \textbf{(\%)} & \textbf{Time (s)} \\ \hline
        Full Model         & \cmark & \cmark & \cmark & \cmark & \cmark & 99.14 & 98.27 & 99.00 & 2046.59 \\
        w/ Std. Attn. + FF & \xmark & \cmark & \cmark & \xmark & \cmark & 98.90 & 97.81 & 98.72 & 1828.79 \\
        w/ Std. Attn. + FF + LN & \xmark & \cmark & \cmark & \xmark & \xmark & 97.15 & 94.72 & 96.66 & 2689.61 \\ \hline
    \end{tabular}}
\end{table}

The ablation study presented in Table \ref{ablation} systematically evaluates the significance of different architectural components of EnergyFormer across three hyperspectral datasets. The results highlight the impact of each component on classification performance and computational efficiency. Across all datasets, the full model consistently achieves the highest OA, AA, and $\kappa$ scores, confirming the effectiveness of integrating MHEA, FoPE, ECBAM, HN, and ELN. Removing MHEA and replacing it with standard self-attention and feed-forward layers (w/ Std. Attn. + FF) results in a performance drop, particularly for the Salinas dataset, where OA declines from 98.73\% to 98.28\%. This underscores the contribution of MHEA in capturing spectral-spatial dependencies more effectively than standard self-attention. Similarly, removing ELN (w/ Std. Attn. + FF + LN) further degrades performance, especially in complex datasets like WHU-Hi-HanChuan, where OA declines significantly from 99.14\% to 97.15\%. The increased computational time in this variant also suggests that ELN contributes to both improved accuracy and efficiency. Overall, these findings validate the design choices of EnergyFormer, demonstrating that its hybrid attention, advanced encoding, and normalization strategies provide a superior trade-off between accuracy and efficiency for HSIC.

\paragraph{\textbf{Effects of Patch Size and Training Samples}} 

EnergyFormer's classification performance is evaluated in terms of patch size and the proportion of training samples. Table \ref{Tab2} presents the impact of varying patch sizes on classification accuracy across three datasets, while Table \ref{tab:ablation_results} examines the model's performance under different training sample percentages. These results provide insights into the model’s sensitivity to spatial context and data availability, influencing both accuracy and training efficiency.

\begin{table}[!hbt]
    \centering
    \caption{Classification performance of EnergyFormer across patch sizes.}
\resizebox{\columnwidth}{!}{\begin{tabular}{c||ccc||ccc||ccc} \hline 
        \multirow{2}{*}{\textbf{Patch}} & \multicolumn{3}{c||}{\textbf{PU}} & \multicolumn{3}{c||}{\textbf{SA}} & \multicolumn{3}{c}{\textbf{HC}} \\ \cline{2-10}
        & \textbf{$\kappa$} & \textbf{OA} & \textbf{AA} & \textbf{$\kappa$} & \textbf{OA} & \textbf{AA} & \textbf{$\kappa$} & \textbf{OA} & \textbf{AA} \\ \hline 
        
        $8 \times 8$  & 97.96 & 98.46 & 97.36 & 96.87 & 97.19 & 98.46 & 97.85 & 98.16 & 96.75 \\
        $10 \times 10$ & 98.13 & 98.59 & 97.73 & 96.75 & 97.08 & 98.54 & 98.45 & 98.67 & 97.41 \\
        $12 \times 12$ & 98.55 & 98.90 & 98.17 & 97.96 & 98.17 & 98.98 & 98.84 & 99.01 & 98.08 \\
        $14 \times 14$ & 98.54 & 98.90 & 98.12 & 98.53 & 98.68 & 99.25 & \textbf{99.13} & \textbf{99.26} & \textbf{98.38} \\
        $16 \times 16$ & 98.87 & 99.15 & 98.58 & 99.10 & 99.19 & 99.62 & 97.58 & 97.93 & 95.87 \\
        $18 \times 18$ & 98.54 & 98.90 & 98.05 & 98.99 & 99.09 & 99.33 & 99.01 & 99.16 & 98.37 \\
        $20 \times 20$ & \textbf{99.05} & \textbf{99.28} & \textbf{98.84} & \textbf{99.10} & \textbf{99.20} & \textbf{99.26} & 99.11 & 99.24 & 98.45 \\ \hline 
    \end{tabular}%
}
\label{Tab2}
\end{table}
\begin{table}[!hbt]
    \centering
    \caption{Classification performance of EnergyFormer across various \% of training samples.}
    \resizebox{\columnwidth}{!}{\begin{tabular}{c|cccc|cccc|cccc} \hline 
        \multirow{2}{*}{Train \%} & \multicolumn{4}{c|}{\textbf{SA}} & \multicolumn{4}{c|}{\textbf{HC}} & \multicolumn{4}{c}{\textbf{PU}} \\ \cline{2-13}
         & \textbf{OA} & \textbf{AA} & \textbf{$\kappa$} & \textbf{Tr Time} & \textbf{OA} & \textbf{AA} & \textbf{$\kappa$} & \textbf{Tr Time} & \textbf{OA} & \textbf{AA} & \textbf{$\kappa$} & \textbf{Tr Time} \\ \hline 
        1\%  &  94.17 &  94.81 &  93.51 &  98.88  &  95.36 &  90.56 &  94.56 &  425.85  &  94.03&  89.25&  92.06&  72.40\\
        3\%  &  97.78 &  98.74 &  97.52 &  258.55 &  98.52 &  97.12 &  98.26 &  1244.58 &  97.95&  96.79&  97.28&  200.39\\
        5\%  &  98.63 &  99.22 &  98.48 &  426.57 &  \textbf{99.28} &  \textbf{98.58} &  \textbf{99.16} &  1233.18 &  98.72&  97.98&  98.31&  347.17\\
        7\%  &  \textbf{99.26} &  \textbf{99.46} &  \textbf{99.18 }&  594.57 &  99.52 &  99.21 &  99.43 &  1717.96 &  \textbf{99.19}&  \textbf{98.49}&  \textbf{98.92}&  459.39\\
        9\%  &  99.55 &  99.76 &  99.50 &  766.92 &  99.65 &  99.31 &  99.59 &  2221.56 &  99.16&  98.24&  98.89&  591.93\\
        10\% &  99.75 &  99.76 &  99.72 &  848.67 &  99.78 &  99.61 &  99.75 &  2447.44 &  99.30&  98.63&  99.08&  657.33\\
        \bottomrule
    \end{tabular}}
    \label{tab:ablation_results}
\end{table}

Table \ref{Tab2} highlights that patch size significantly affects classification performance, with an optimal range observed between $14 \times 14$ and $20 \times 20$, where the model achieves peak accuracy. The results indicate that larger patches provide richer spatial context, but excessive enlargement may lead to performance degradation due to redundant information. Meanwhile, Table \ref{tab:ablation_results} demonstrates that increasing training samples improves classification accuracy, particularly for complex datasets like UH, where performance stabilizes beyond 7\% of training samples. However, this comes at the cost of increased training time, underscoring the trade-off between model accuracy and computational efficiency.

\paragraph{\textbf{Comparison with CNN, Transformer, and Mamba Models}}

The effectiveness of HSIC hinges on the ability to capture spatial and spectral dependencies efficiently. CNNs, Transformer-based architectures, and Mamba-based state space models have each contributed to advancing performance in this domain. However, their inherent limitations in modeling long-range dependencies, adaptively prioritizing spectral features, and maintaining computational efficiency necessitate further innovation. To this end, we introduce EnergyFormer (EF), a novel model that leverages a refined spectral-spatial representation strategy, outperforming existing approaches across multiple datasets. The following comparative evaluation demonstrates its superiority against leading methods, including Hybrid CNN (HCNN) \cite{roy2019hybridsn}, Multihead spatial-spectral mamba (MHM) \cite{ahmad2024}, Attention graph CNN (AGC) \cite{10409250}, pyramid spatial-spectral transformer (PF) \cite{10681622}, Morphological spatial-spectral transformer (MF) \cite{roy2023spectral}, spatial-spectral Former (SF) \cite{10604879}, and wavelet-based spatial-spectral mamba (WM) \cite{10767233}.

\begin{table}[!hbt]
    \centering
    \caption{Performance comparison on the \textbf{HC dataset} across class-wise accuracies and aggregate metrics.}
    \resizebox{\columnwidth}{!}{%
    \begin{tabular}{c|cccccccc} \hline
        \textbf{Class} & AGC & HCNN& MF & PF & SF & MHM & WM & \textbf{\textit{EF}} \\ \hline
        Strawberry    & 97.89& 98.63& 98.32& 97.89& 99.44& 97.68& 97.35& 99.67\\
        Cowpea        & 93.14& 96.58& 96.02& 96.52& 98.20& 84.04& 89.73& 98.90\\
        Soybean       & 98.98& 98.87& 97.47& 95.04& 98.63& 92.11& 90.63& 99.65\\
        Sorghum       & 99.58& 99.19& 97.51& 98.83& 98.72& 96.99& 95.86& 99.69\\
        Water spinach & 85.28& 99.91& 92.97& 68.14& 93.50& 83.79& 87.03& 99.91\\
        Watermelon    & 70.47& 80.39& 78.90& 82.40& 85.50& 59.14& 70.39& 96.94\\
        Greens        & 97.76& 95.46& 96.42& 93.33& 95.59& 92.79& 92.37& 98.76\\
        Trees         & 95.12& 95.38& 94.15& 91.73& 95.41& 80.19& 86.00& 99.17\\
        Grass         & 93.90& 96.94& 95.19& 93.07& 96.49& 81.14& 83.32& 98.38\\
        Red roof      & 99.03& 99.27& 98.39& 99.41& 98.98& 93.45& 98.71& 99.54\\
        Gray roof     & 97.52& 97.16& 98.72& 95.70& 99.05& 93.96& 96.82& 98.86\\
        Plastic       & 96.95& 97.36& 94.93& 93.75& 97.42& 35.48& 75.17& 98.55\\
        Bare soil     & 89.37& 88.39& 90.26& 83.70& 89.23& 69.66& 77.00& 96.62\\
        Road          & 94.98& 97.37& 97.39& 95.31& 96.78& 89.00& 92.31& 99.25\\
        Bright object & 71.14& 92.30& 85.63& 85.82& 83.13& 83.65& 65.85& 93.45\\
        Water         & 99.85& 99.35& 99.88& 99.70& 99.90& 89.73& 99.51& 99.91\\ \hline
        
        \textbf{$\kappa$} & 96.04& 97.20& 96.82& 95.56& 97.63& 89.31& 92.45& \textbf{99.16}\\ \hline
        \textbf{OA}  & 96.62& 97.61& 97.28& 96.20& 97.98& 90.88& 93.55& \textbf{99.28}\\ \hline
        \textbf{AA}  & 92.56& 95.83& 94.51& 91.90& 95.38& 83.24& 87.38& \textbf{98.58}\\ \hline
    \end{tabular}}
    \label{Tab3}
\end{table}
\begin{figure}[!hbt]
    \centering
    \begin{subfigure}{0.05\textwidth}
	\includegraphics[width=0.99\textwidth]{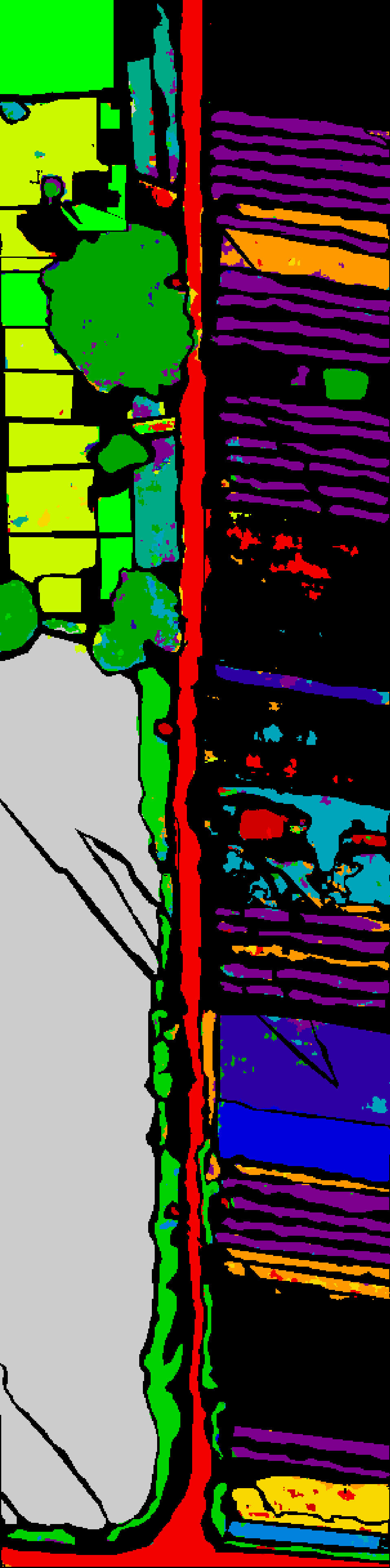}
	\caption*{AGC} 
    \end{subfigure}
     \begin{subfigure}{0.05\textwidth}
	\includegraphics[width=0.99\textwidth]{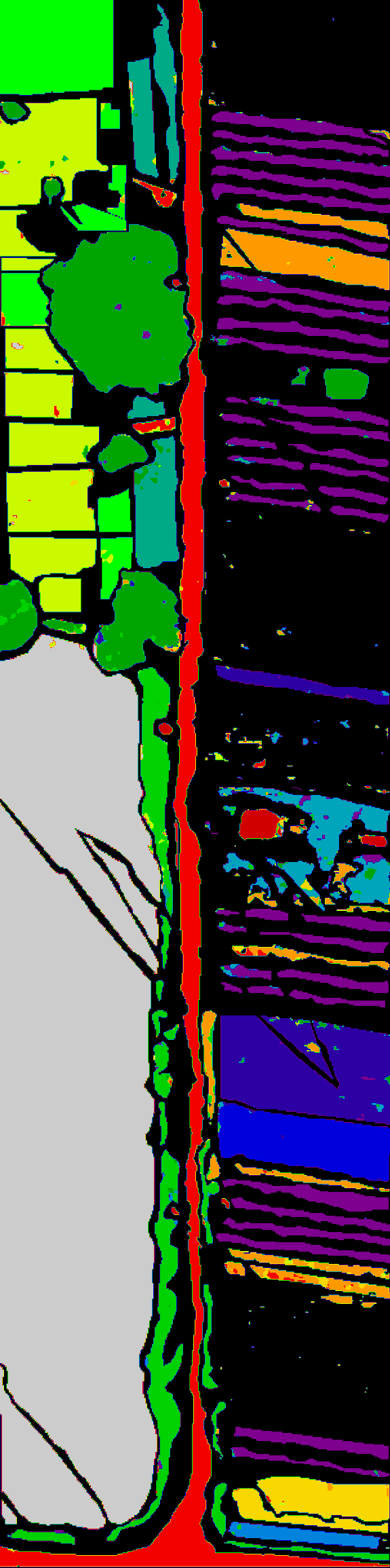}
	\caption*{HCNN}
    \end{subfigure}
     \begin{subfigure}{0.05\textwidth}
	\includegraphics[width=0.99\textwidth]{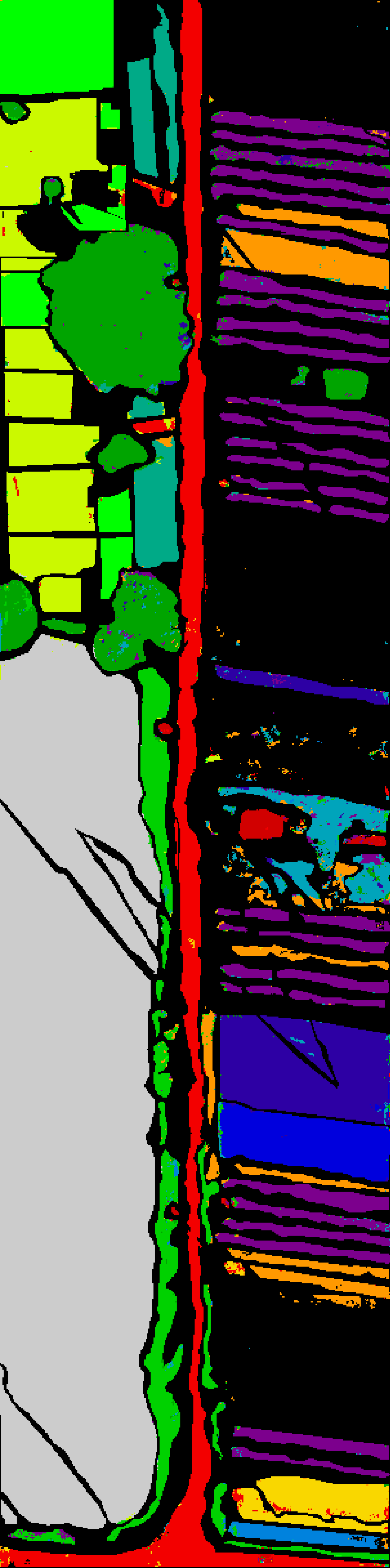}
	\caption*{MF}
    \end{subfigure}
    \begin{subfigure}{0.05\textwidth}
	\includegraphics[width=0.99\textwidth]{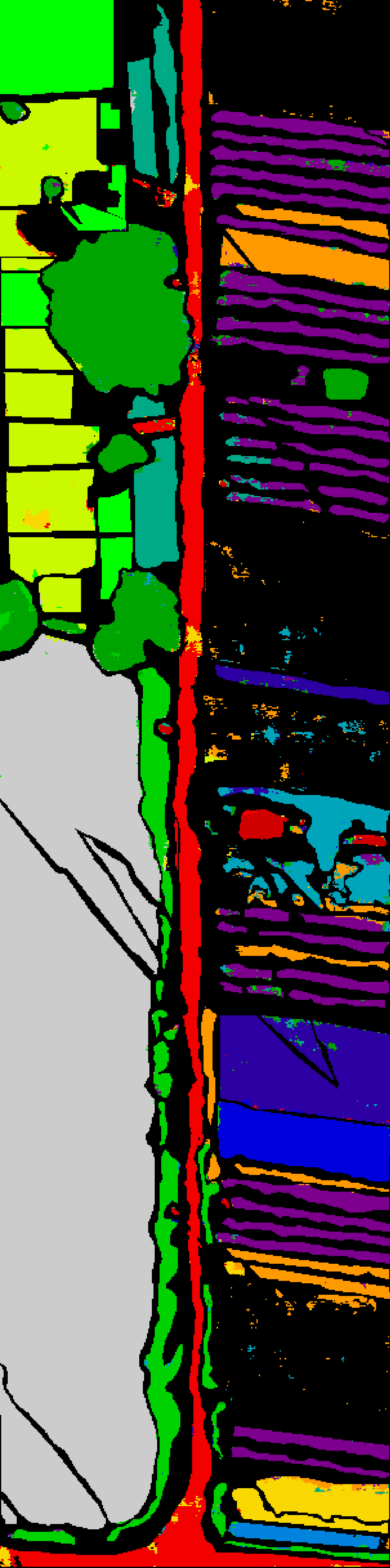}
	\caption*{PF}
    \end{subfigure} 
    \begin{subfigure}{0.05\textwidth}
	\includegraphics[width=0.99\textwidth]{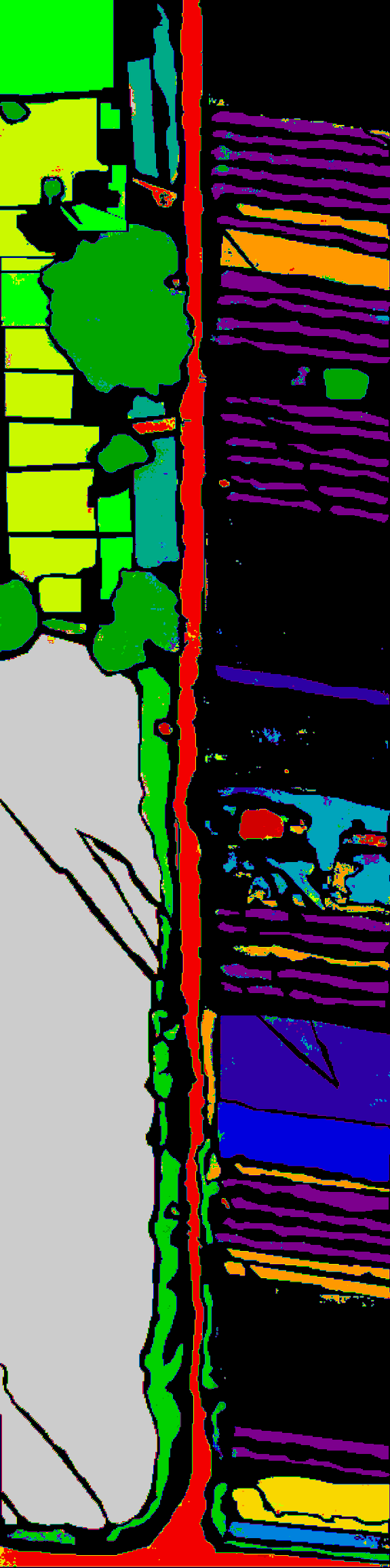}
	\caption*{SF}
    \end{subfigure}
    \begin{subfigure}{0.05\textwidth}
	\includegraphics[width=0.99\textwidth]{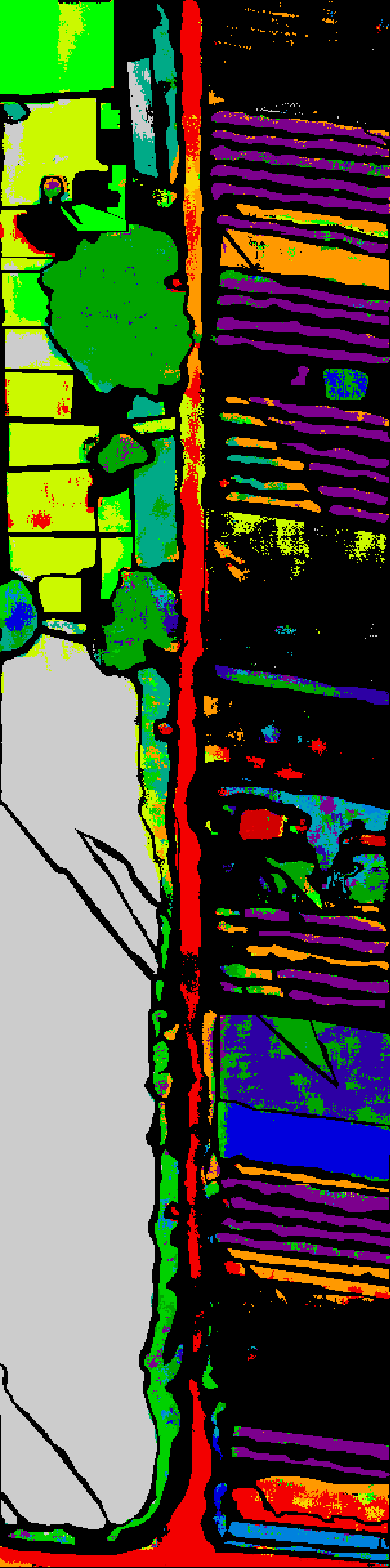}
	\caption*{MHM}
    \end{subfigure}
    \begin{subfigure}{0.05\textwidth}
	\includegraphics[width=0.99\textwidth]{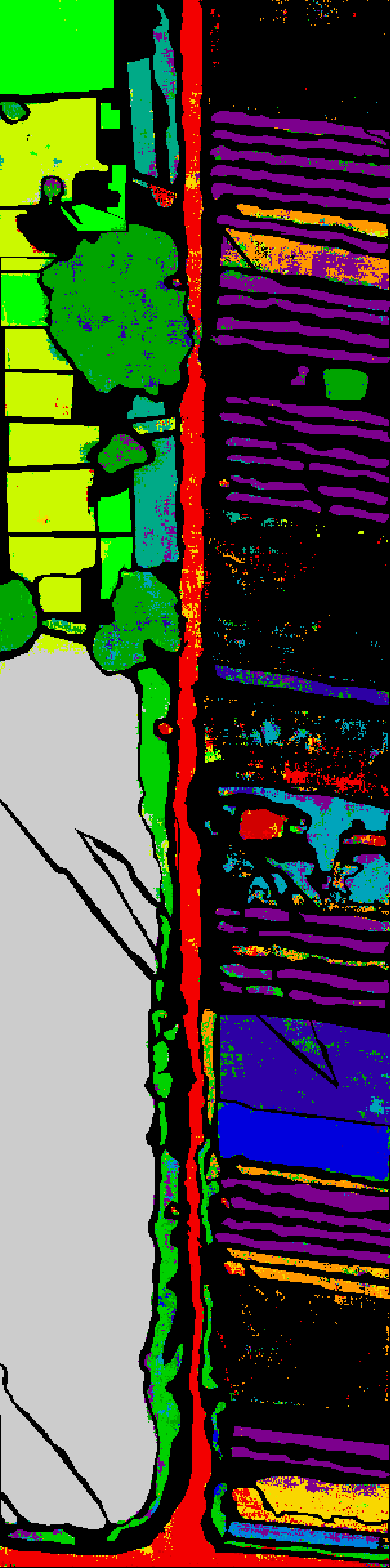}
	\caption*{WM}
    \end{subfigure}
    \begin{subfigure}{0.05\textwidth}
	\includegraphics[width=0.99\textwidth]{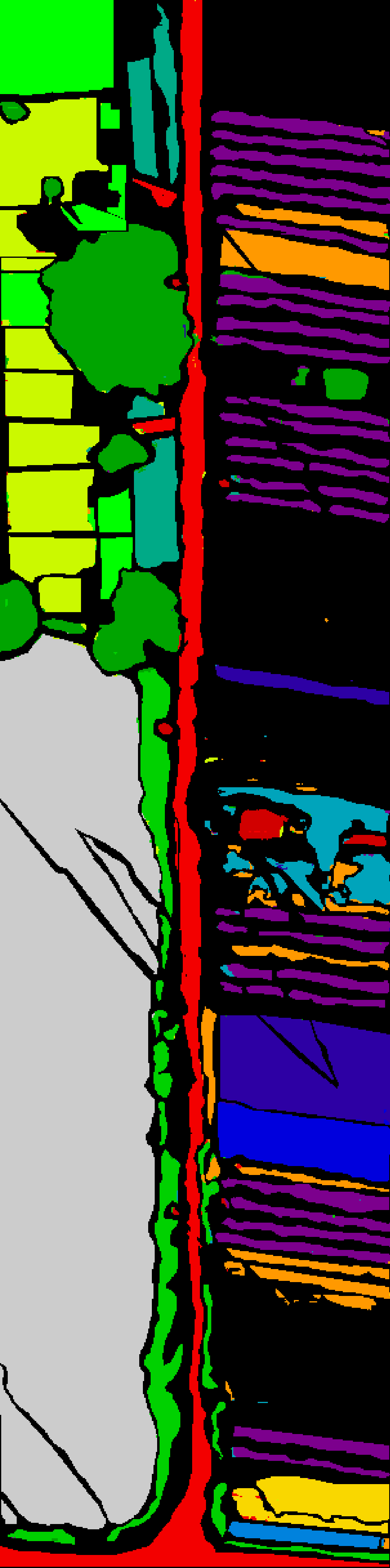}
	\caption*{EF}
    \end{subfigure}
\caption{Classification maps for the \textbf{HC dataset}, highlighting spatial variability and class-specific performance.}
\label{Fig7}
\end{figure}
\begin{table}[!hbt]
    \centering
    \caption{Performance comparison on the \textbf{SA dataset} across class-wise accuracies and aggregate metrics.}
    \resizebox{\columnwidth}{!}{%
    \begin{tabular}{c|cccccccc} \hline
        \textbf{Class} & AGC & HCNN& MF & PF & SF & MHM & WM & \textbf{\textit{EF}} \\ \hline 
        Broccoli 1        & 97.79& 100.0 & 98.51& 99.94& 100.0& 99.77& 99.72& 100.0\\
        Broccoli 2        & 100.0& 100.0 & 100.0& 100.0& 100.0& 99.85& 99.94& 99.97\\
        Fallow            & 97.30& 80.44 & 98.71& 100.0& 99.09& 99.76& 99.43& 99.78\\
        Fallow Rough      & 99.68& 99.54& 99.28& 99.20& 99.54& 96.97& 98.08& 97.29\\
        Fallow Smooth     & 96.39& 99.80& 95.85& 99.33& 99.41& 58.00& 97.46& 99.98\\
        Stubble           & 99.92& 99.97& 100.0& 99.88& 100.0& 100.0& 99.91& 99.97\\
        Celery            & 100.0& 100.0& 99.47& 99.90& 99.97& 99.84& 99.16& 99.57\\
        Grapes            & 97.21& 97.77& 91.49& 98.01& 94.10& 96.04& 96.94& 96.16\\
        Soil Vinyard      & 99.91& 100.0& 99.73& 100.0& 99.77& 99.55& 99.96& 100.0\\
        Corn Senesced     & 94.27& 99.13& 97.83& 78.07& 97.43& 92.88& 98.91& 98.48\\
        Lettuce 4wk       & 99.27& 96.94& 92.00& 99.79& 96.94& 99.16& 99.27& 100.0\\
        Lettuce 5wk       & 90.31& 100.0& 99.08& 99.82& 99.94& 99.01& 99.94& 99.77\\
        Lettuce 6wk       & 100.0& 98.04& 97.45& 99.03& 98.85& 98.18& 97.45& 99.52\\
        Lettuce 7wk       & 96.26& 97.34& 92.43& 99.79& 100.0& 98.13& 97.61& 99.69\\
        Vinyard Untrained & 97.22& 91.65& 89.15& 98.18& 88.52& 55.52& 95.16& 97.51\\
        Vinyard Vertical  & 98.40& 98.83& 96.93& 98.52& 98.13& 98.70& 98.64& 99.94\\ \hline 
        
        \textbf{$\kappa$} & 97.58& 97.15& 95.28& 97.62& 96.47& 89.19& 98.02& \textbf{98.48}\\ \hline
        \textbf{OA}      & 97.83& 97.44& 95.76& 97.87& 96.83& 90.33& 98.22& \textbf{98.63}\\ \hline
        \textbf{AA}      & 97.75& 97.47& 96.74& 98.10& 98.23& 93.15& 98.60& \textbf{99.22}\\ \hline
    \end{tabular}}
    \label{Tab6}
\end{table}
\begin{figure}[!hbt]
    \centering
    \begin{subfigure}{0.05\textwidth}
	\includegraphics[width=0.99\textwidth]{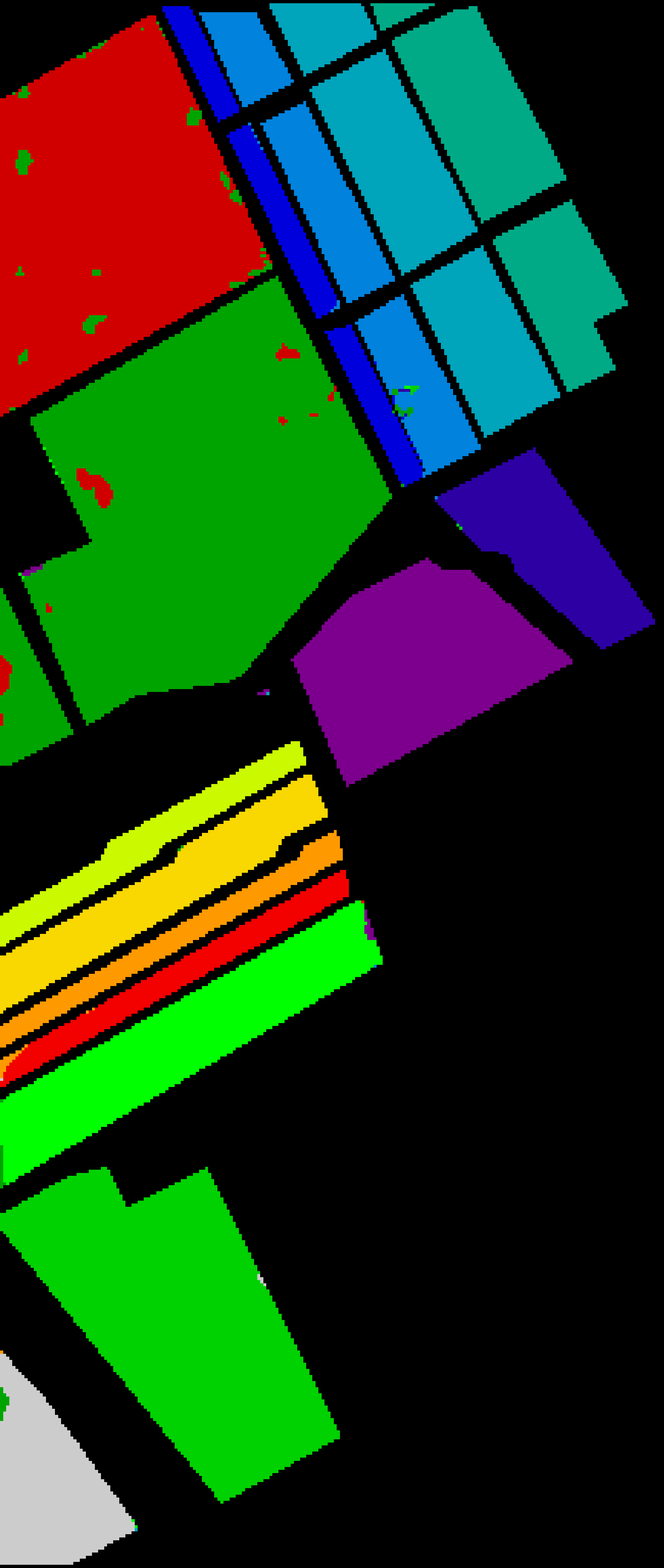}
	\caption*{AGC} 
    \end{subfigure}
     \begin{subfigure}{0.05\textwidth}
	\includegraphics[width=0.99\textwidth]{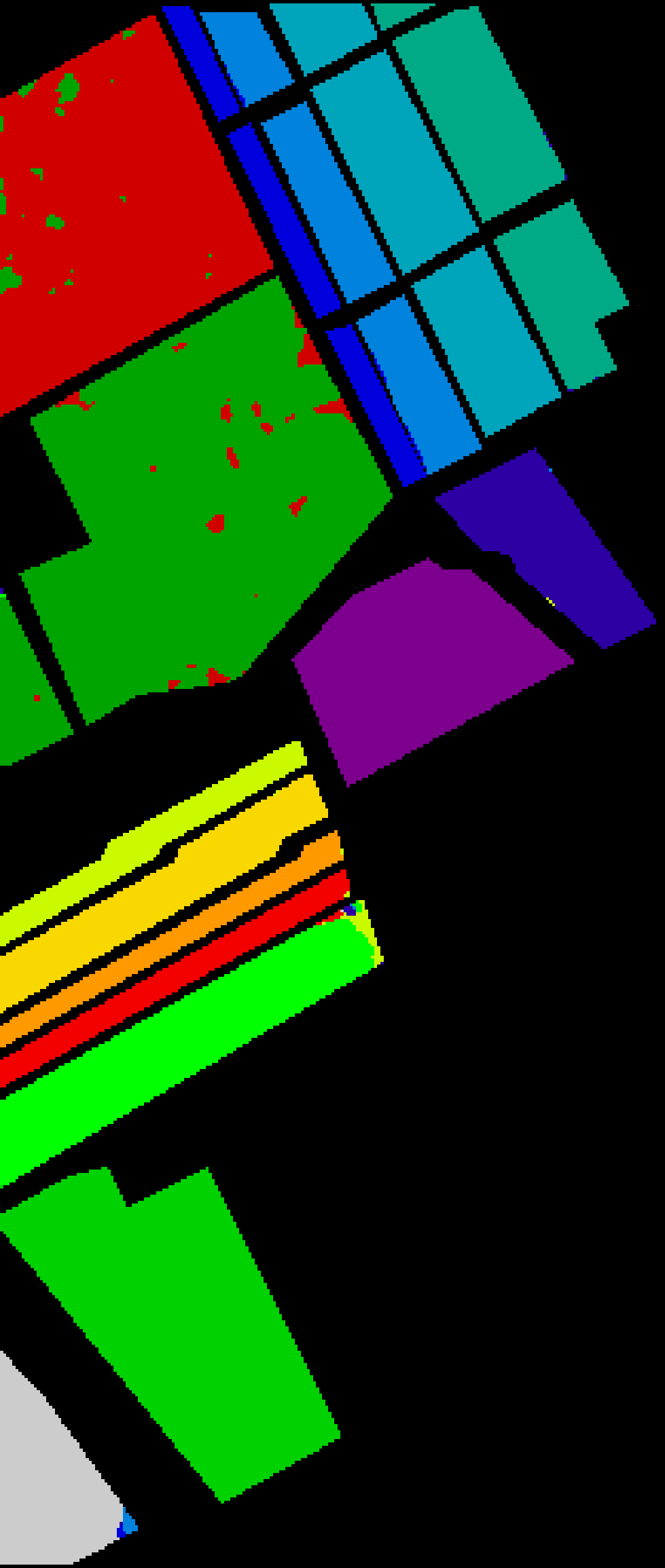}
	\caption*{HCNN}
    \end{subfigure}
     \begin{subfigure}{0.05\textwidth}
	\includegraphics[width=0.99\textwidth]{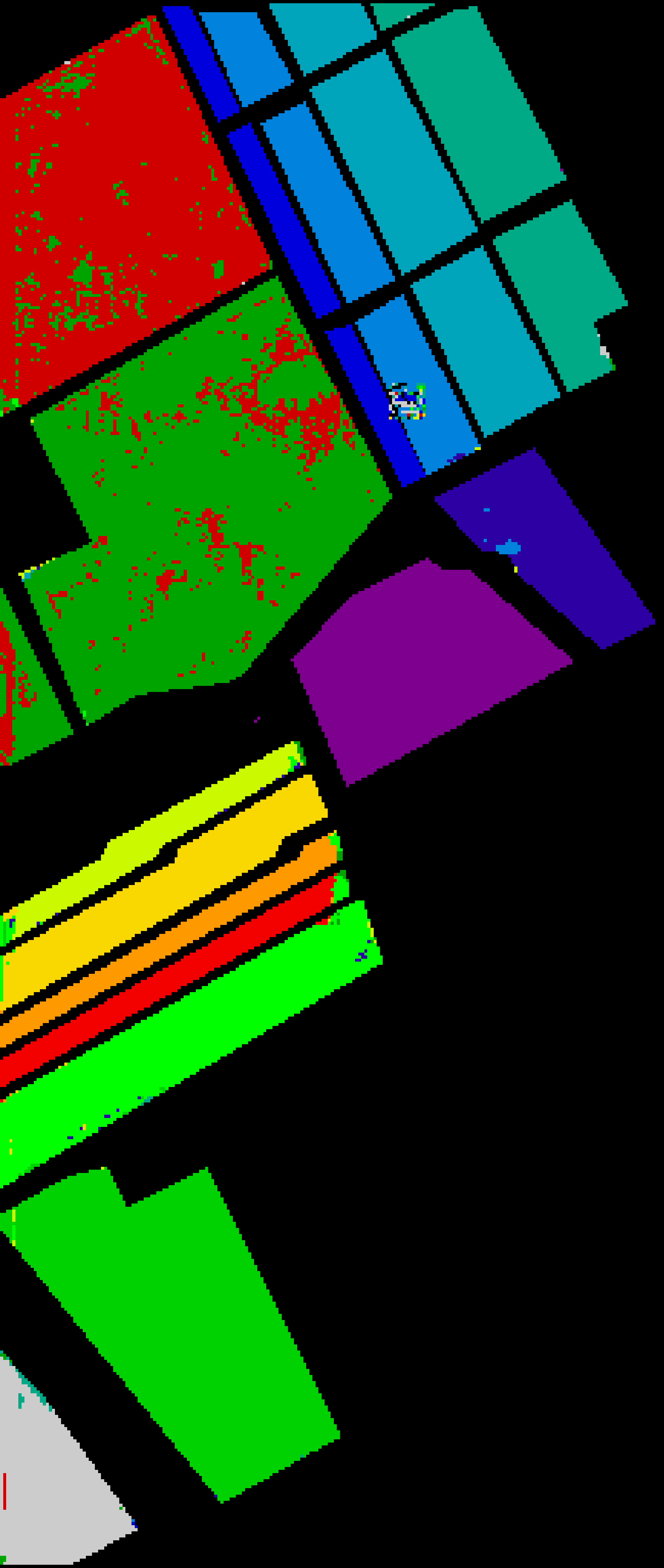}
	\caption*{MF}
    \end{subfigure}
    \begin{subfigure}{0.05\textwidth}
	\includegraphics[width=0.99\textwidth]{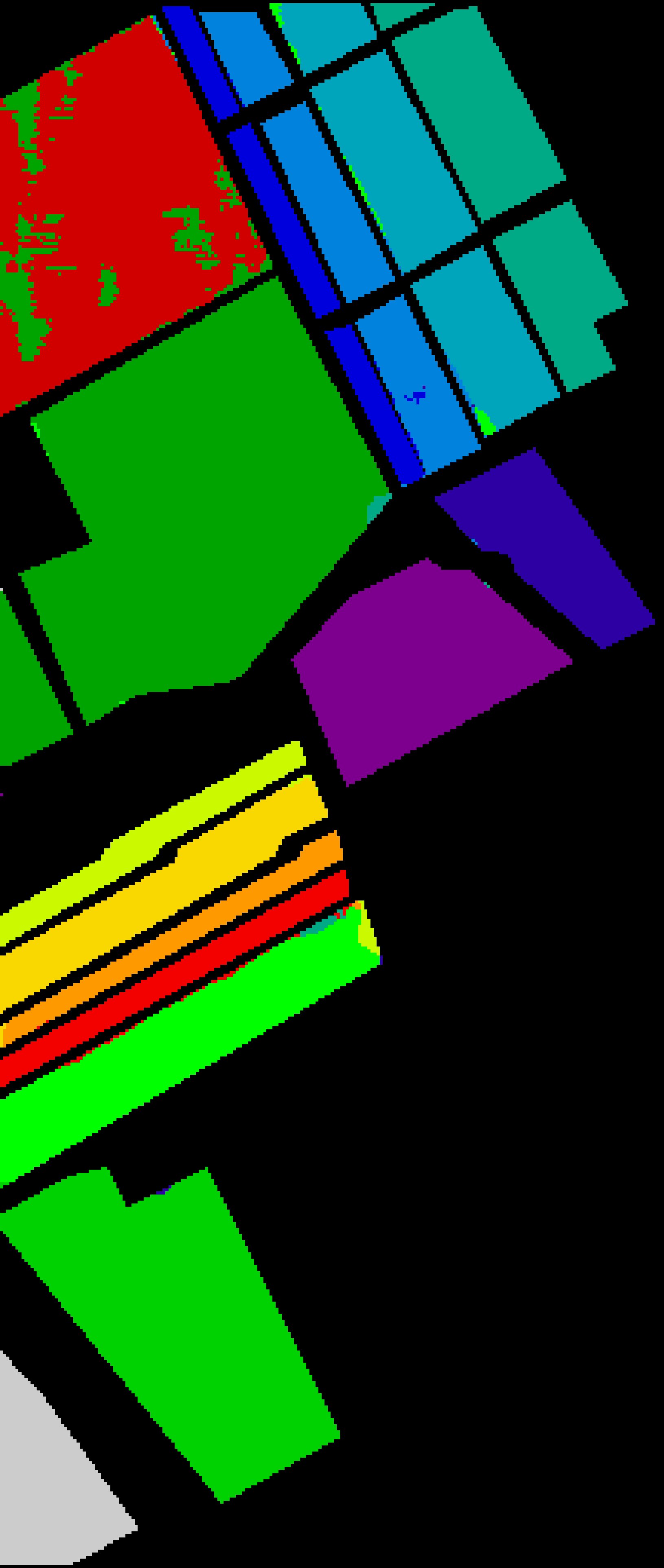}
	\caption*{PF}
    \end{subfigure} 
    \begin{subfigure}{0.05\textwidth}
	\includegraphics[width=0.99\textwidth]{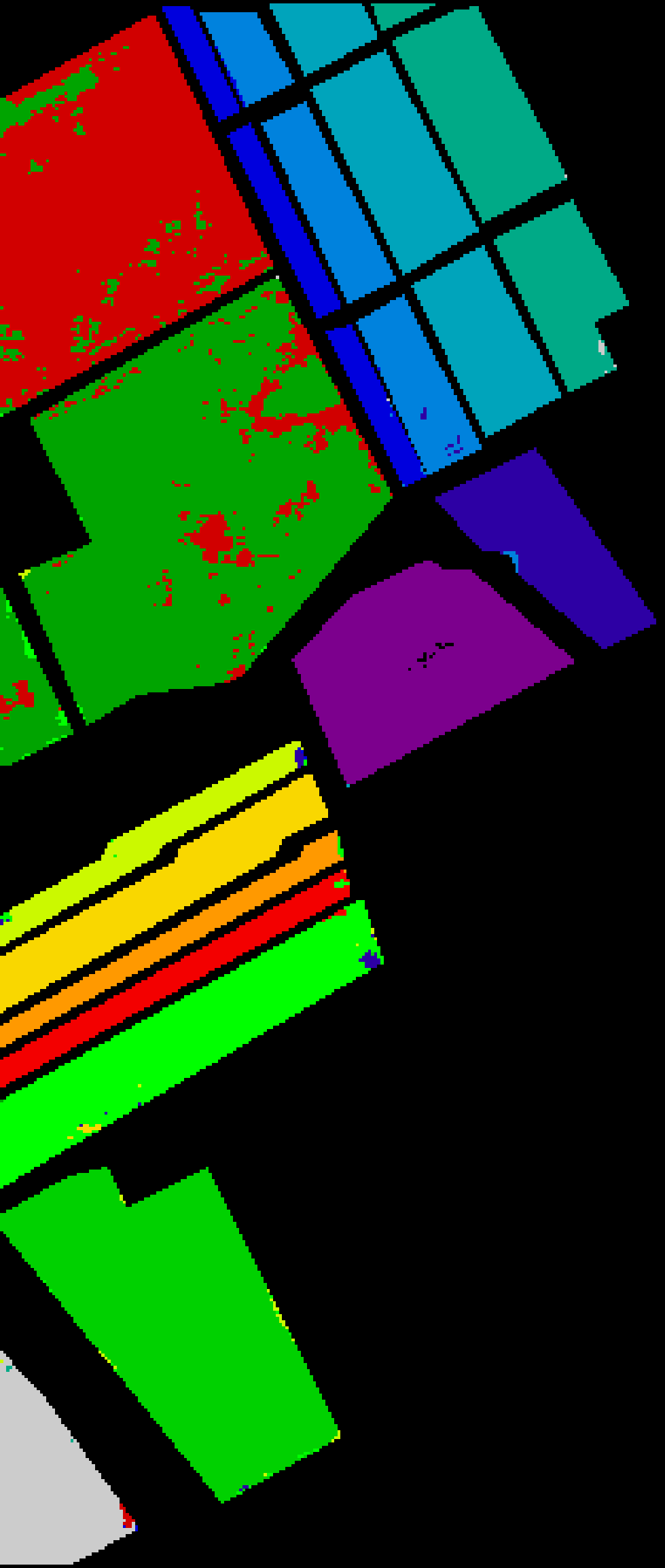}
	\caption*{SF}
    \end{subfigure}
    \begin{subfigure}{0.05\textwidth}
	\includegraphics[width=0.99\textwidth]{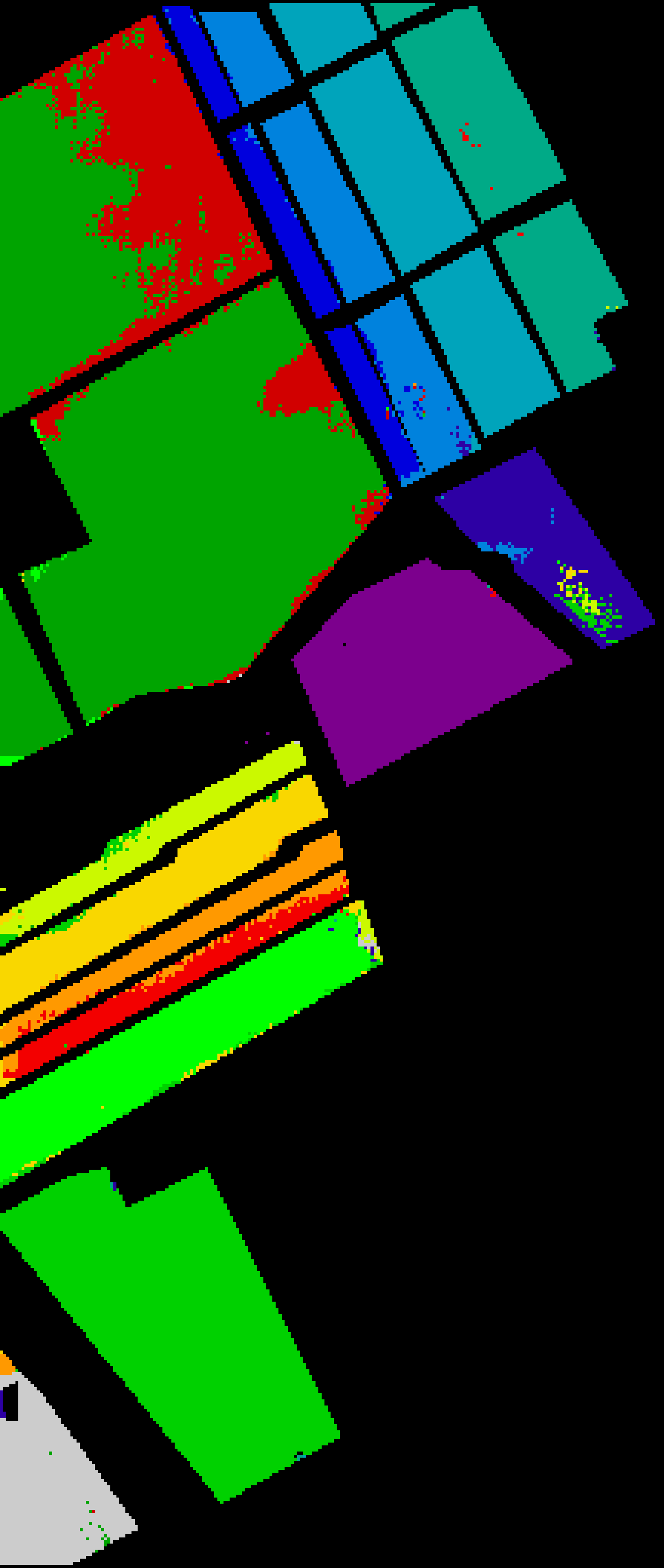}
	\caption*{MHM}
    \end{subfigure}
    \begin{subfigure}{0.05\textwidth}
	\includegraphics[width=0.99\textwidth]{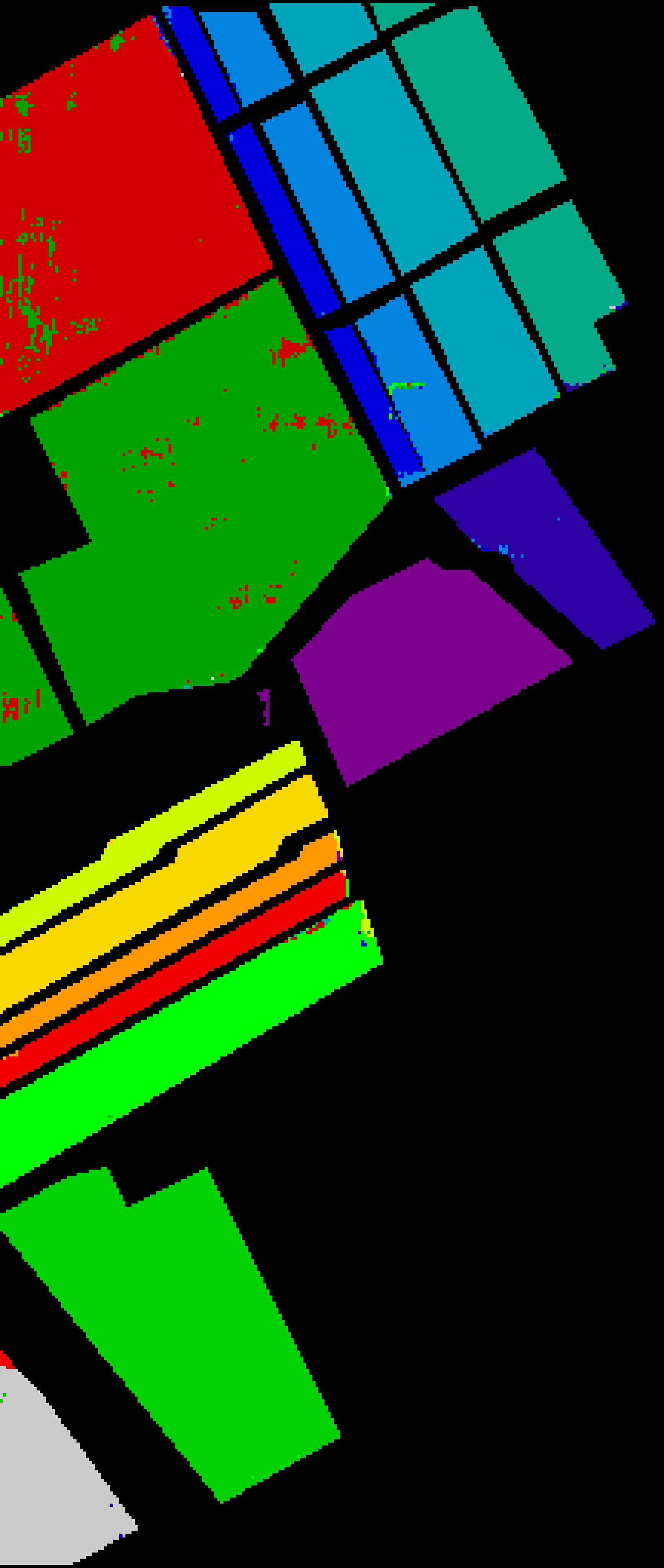}
	\caption*{WM}
    \end{subfigure}
    \begin{subfigure}{0.05\textwidth}
	\includegraphics[width=0.99\textwidth]{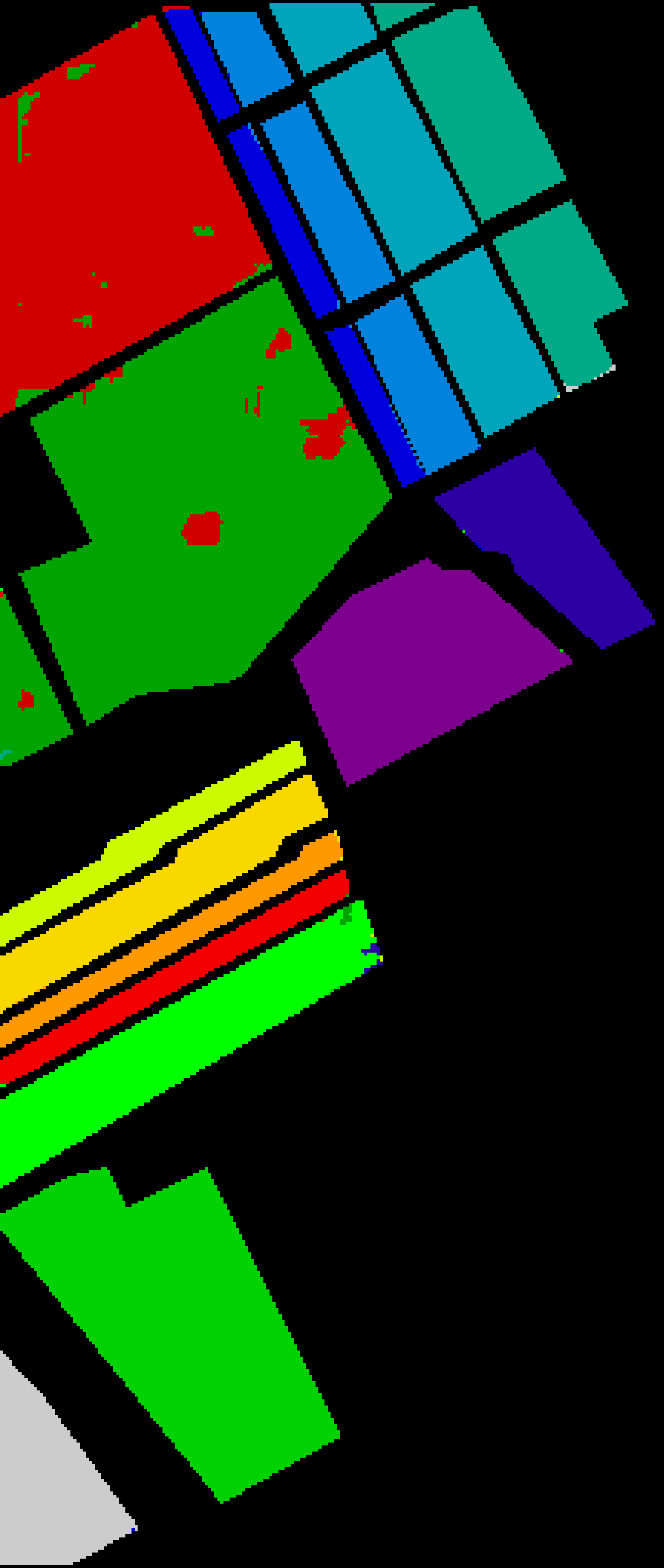}
	\caption*{EF}
    \end{subfigure}
\caption{Classification maps for the \textbf{SA dataset}, highlighting spatial variability and class-specific performance.}
\label{Fig8}
\end{figure}
\begin{table}[!hbt]
    \centering
    \caption{Performance comparison on the \textbf{PU dataset} across class-wise accuracies and aggregate metrics.}
    \resizebox{\columnwidth}{!}{\begin{tabular}{c|cccccccc} \hline
        \textbf{Class} & AGC & HCNN& MF & PF & SF & MHM & WM & \textit{EF} \\ \hline 
        Asphalt & 95.01& 98.58& 94.29& 98.45& 95.57& 82.89& 97.13& 98.76\\
        Meadows & 99.85& 99.90& 97.67& 99.79& 99.93& 99.02& 98.86& 99.56\\
        Gravel & 83.22& 94.48& 75.56& 96.71& 88.21& 68.39& 84.70& 98.04\\
        Trees & 96.48& 97.90& 97.10& 98.29& 97.73& 82.74& 94.81& 95.90\\
        Metal Sheets & 98.60& 99.84& 99.26& 99.66& 100.0& 96.77& 99.50& 99.59\\
        Bare Soil & 99.09& 97.71& 89.57& 97.61& 90.57& 93.43& 95.47& 99.47\\
        Bitumen & 82.62& 97.70& 76.29& 98.07& 84.00& 84.46& 90.39& 97.83\\
        Bricks & 95.93& 94.05& 91.31& 80.65& 89.47& 75.28& 88.92& 96.89\\
        Shadows & 90.26& 100.0& 98.71& 99.06& 99.77& 28.52& 79.57& 95.78\\ \hline 
        \textbf{$\kappa$} & 95.79& 98.02& 91.94& 96.49& 94.71& 85.41& 94.27& \textbf{98.31}\\ \hline
        \textbf{OA} & 96.83& 98.51& 93.93& 97.35& 96.03& 89.06& 95.68& \textbf{98.72}\\ \hline
        \textbf{AA} & 93.45& 97.83& 91.08& 96.48& 93.92& 79.05& 92.15& \textbf{97.98}\\ \hline
    \end{tabular}}
    \label{Tab5}
\end{table}
\begin{figure}[!hbt]
    \centering
    \begin{subfigure}{0.05\textwidth}
	\includegraphics[width=0.99\textwidth]{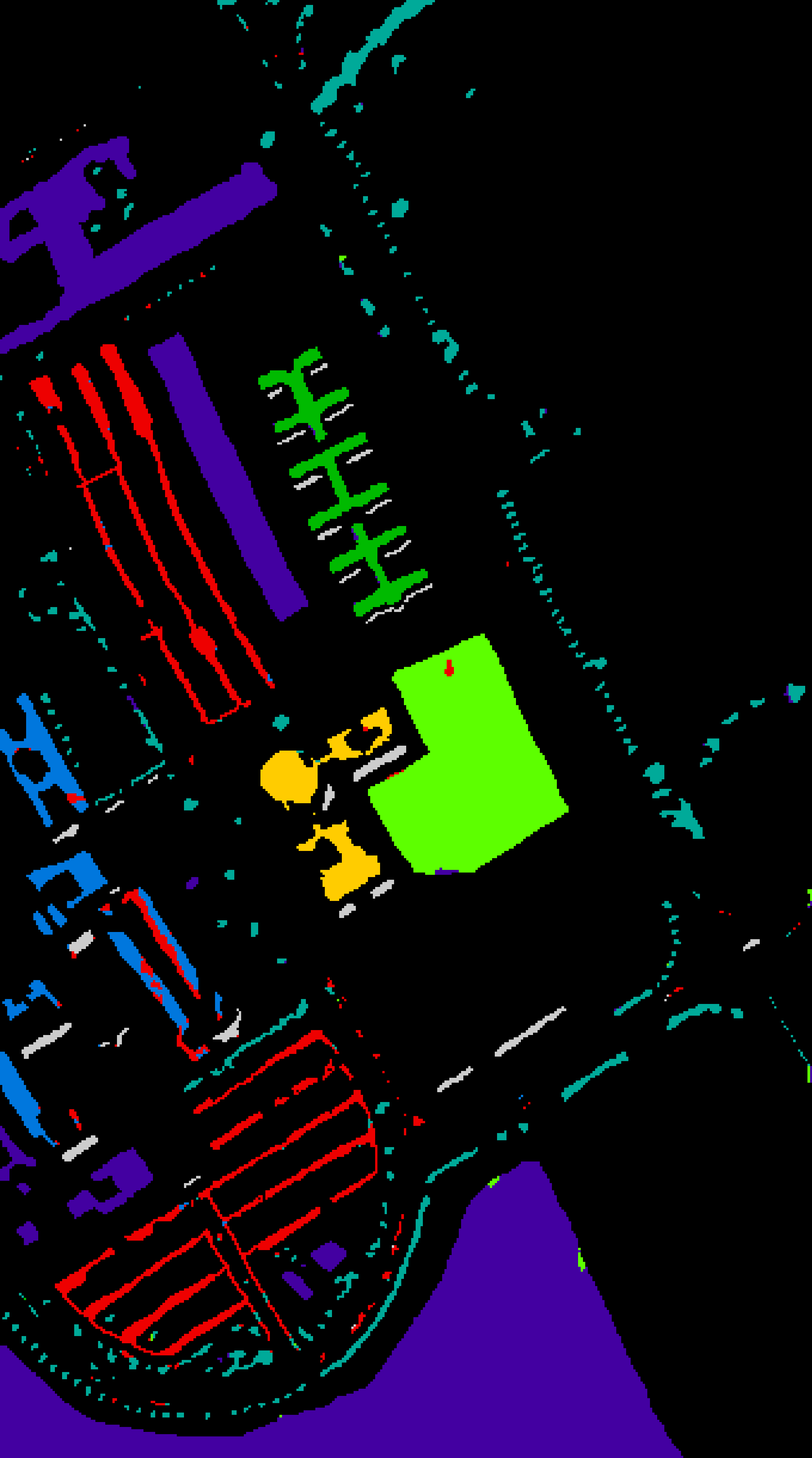}
	\caption*{AGC} 
    \end{subfigure}
     \begin{subfigure}{0.05\textwidth}
	\includegraphics[width=0.99\textwidth]{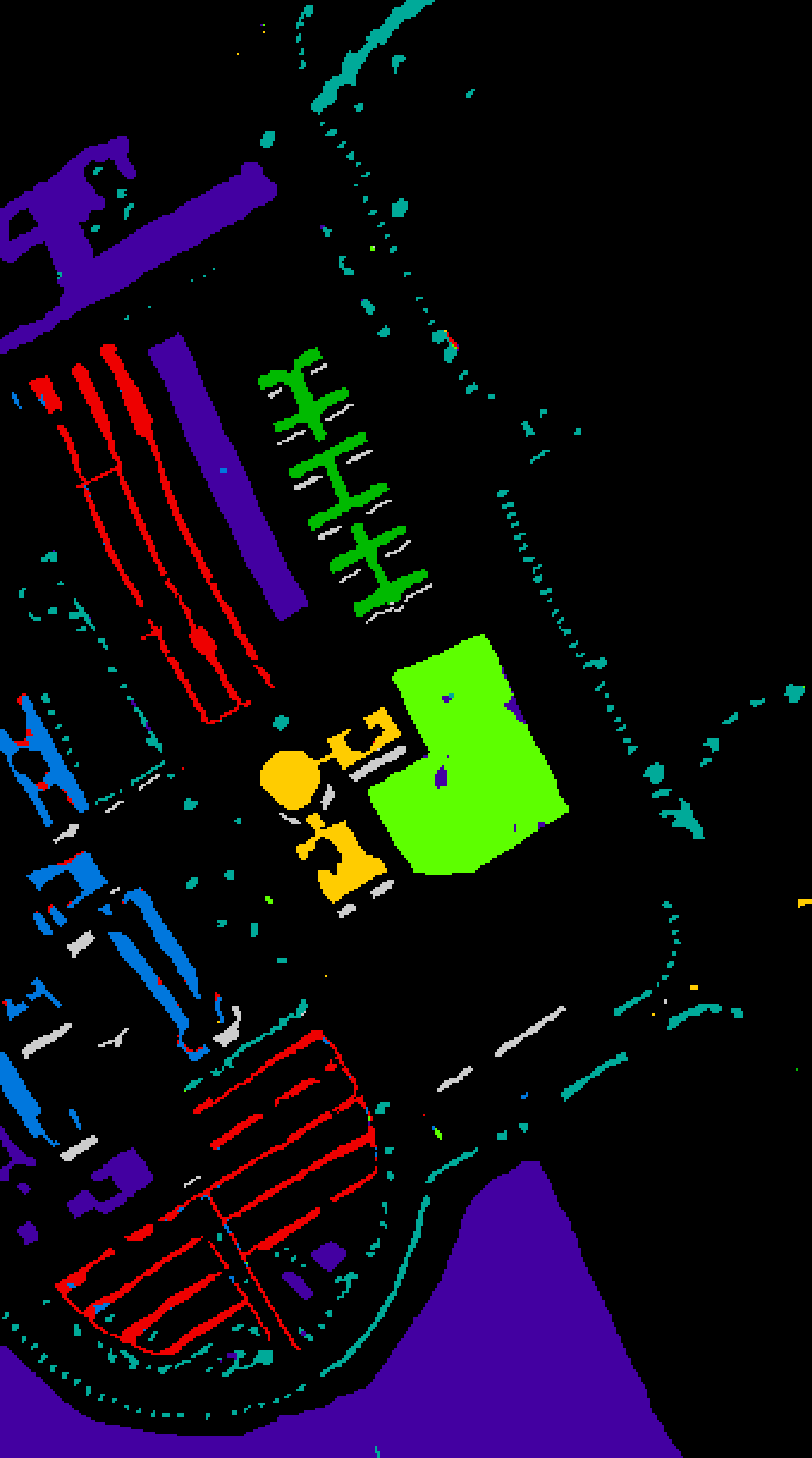}
	\caption*{HCNN}
    \end{subfigure}
     \begin{subfigure}{0.05\textwidth}
	\includegraphics[width=0.99\textwidth]{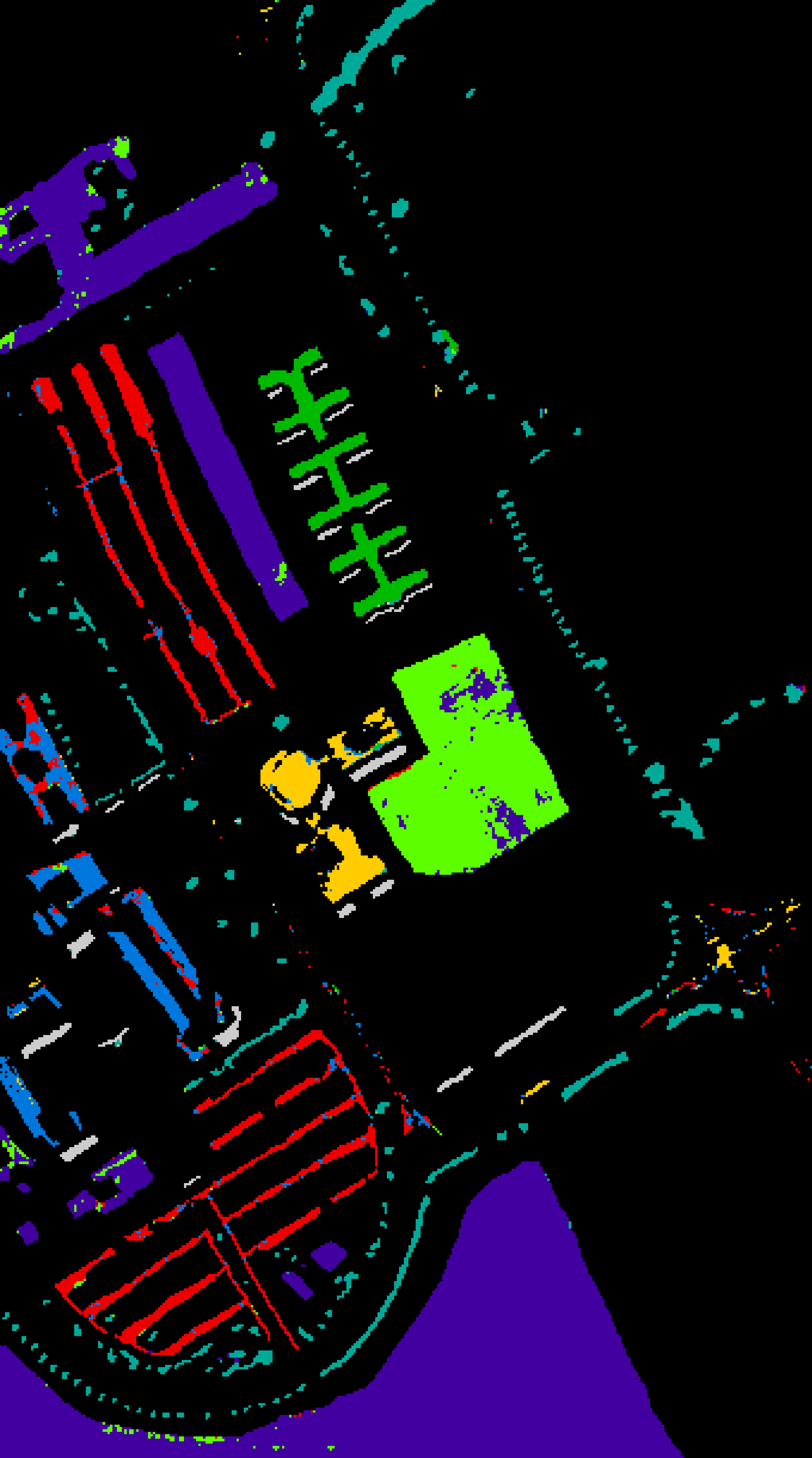}
	\caption*{MF}
    \end{subfigure}
    \begin{subfigure}{0.05\textwidth}
	\includegraphics[width=0.99\textwidth]{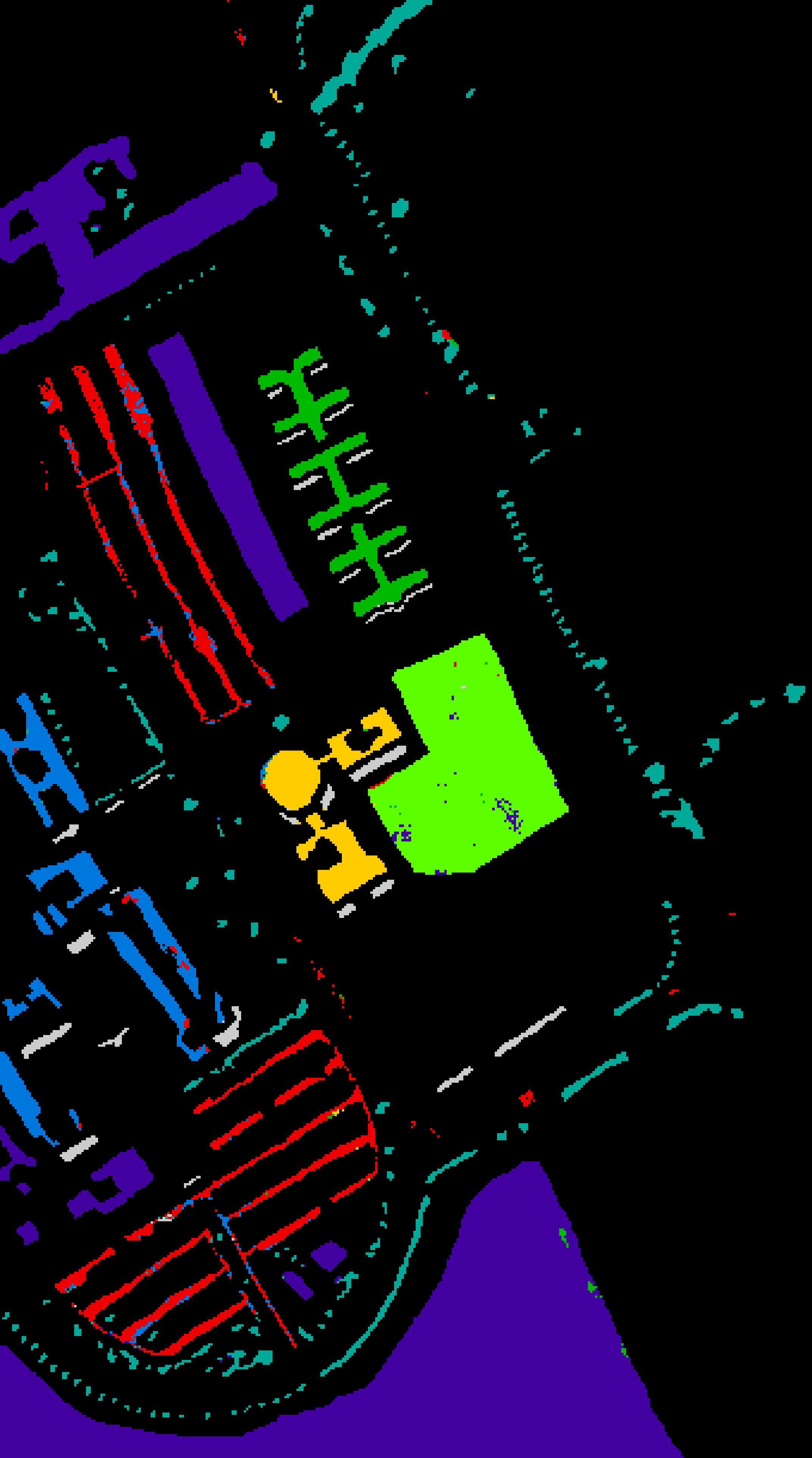}
	\caption*{PF}
    \end{subfigure} 
    \begin{subfigure}{0.05\textwidth}
	\includegraphics[width=0.99\textwidth]{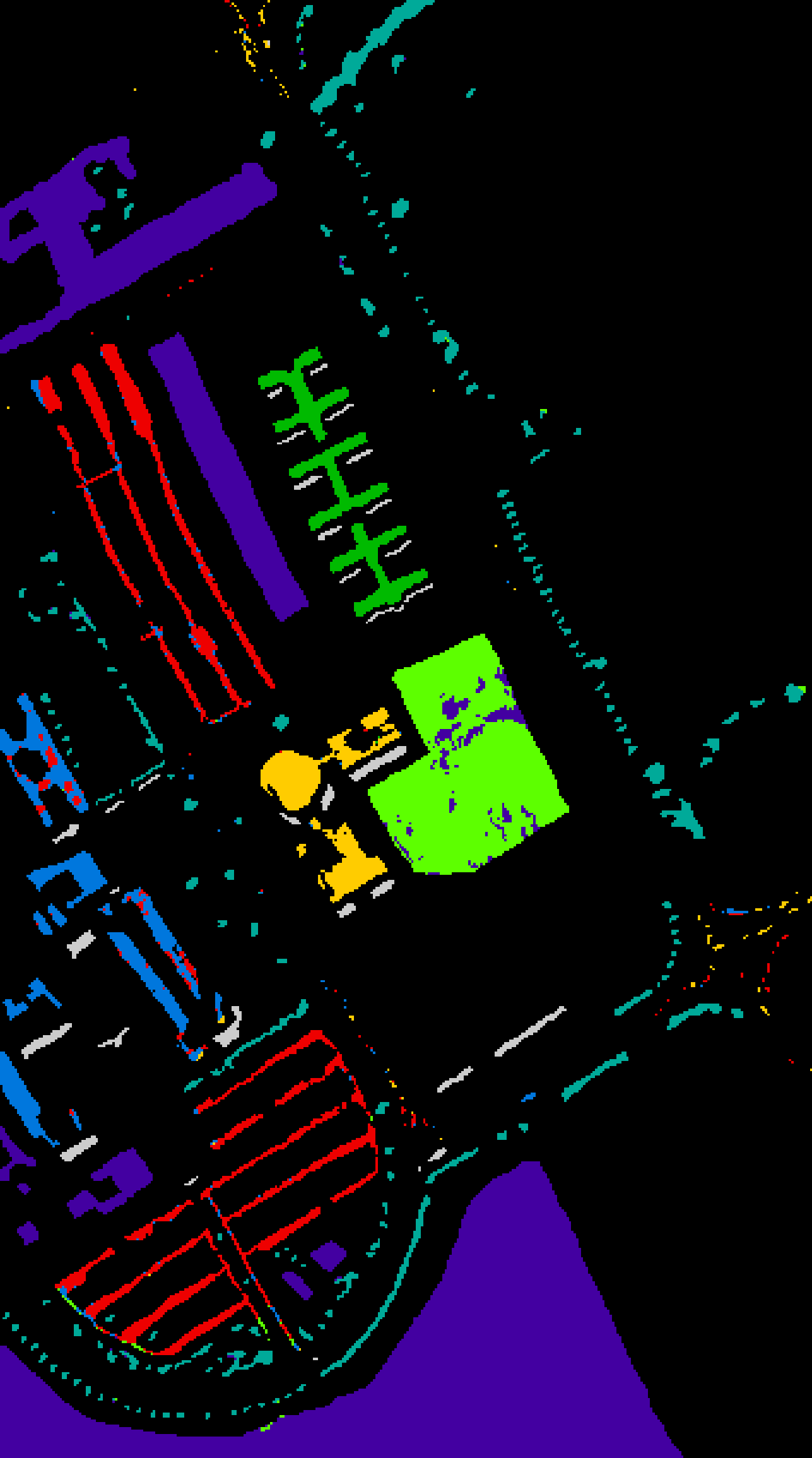}
	\caption*{SF}
    \end{subfigure}
    \begin{subfigure}{0.05\textwidth}
	\includegraphics[width=0.99\textwidth]{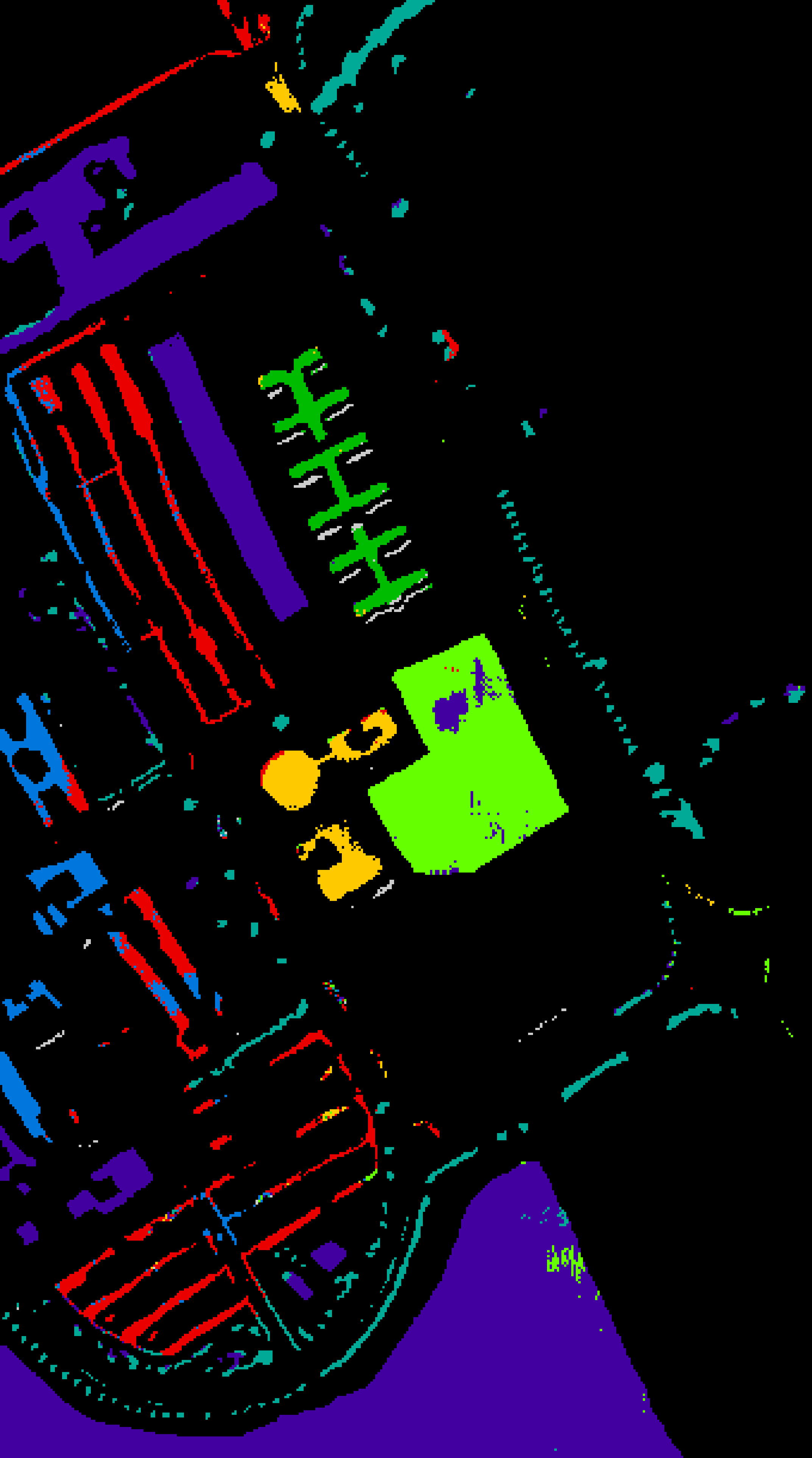}
	\caption*{MHM}
    \end{subfigure}
    \begin{subfigure}{0.05\textwidth}
	\includegraphics[width=0.99\textwidth]{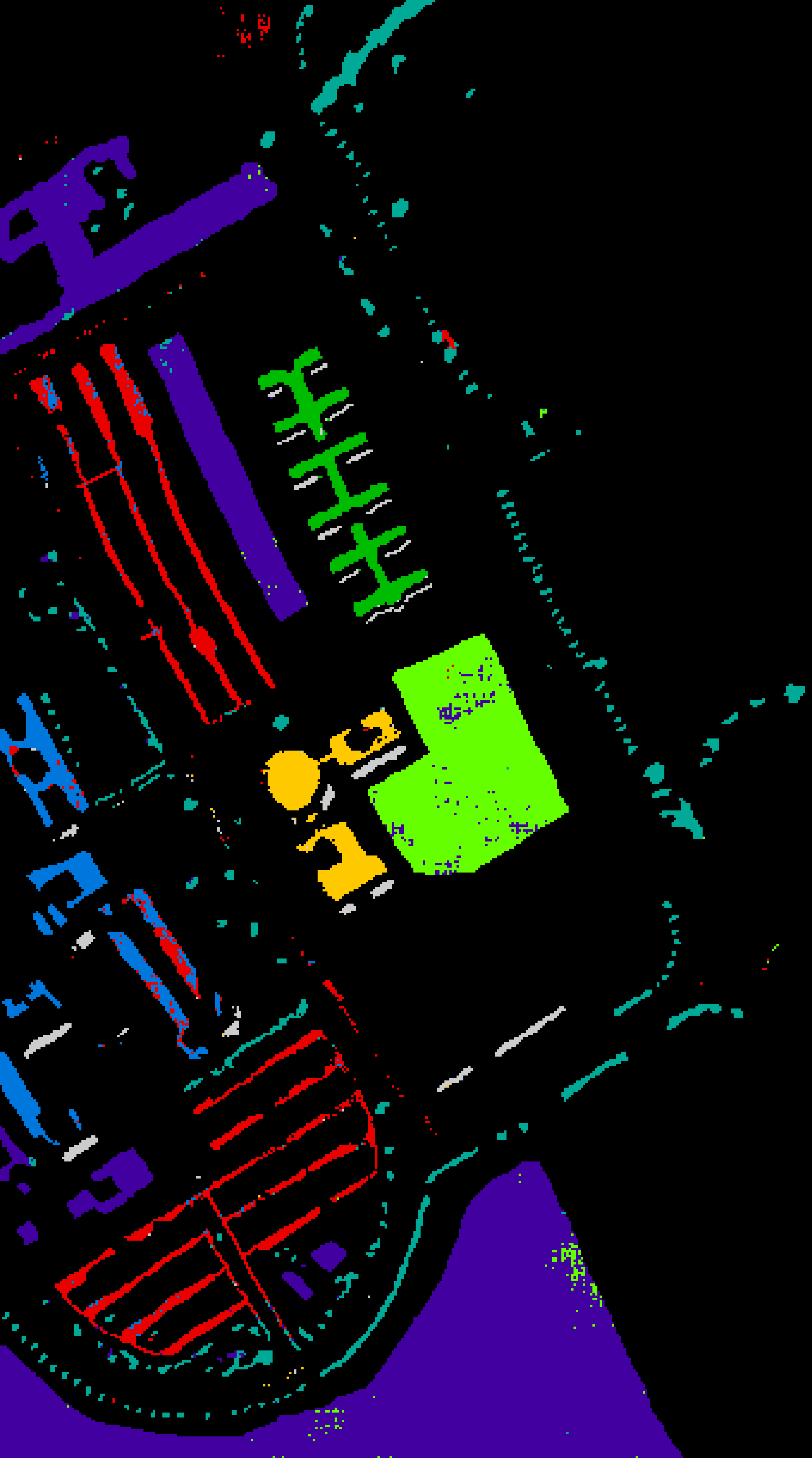}
	\caption*{WM}
    \end{subfigure}
    \begin{subfigure}{0.05\textwidth}
	\includegraphics[width=0.99\textwidth]{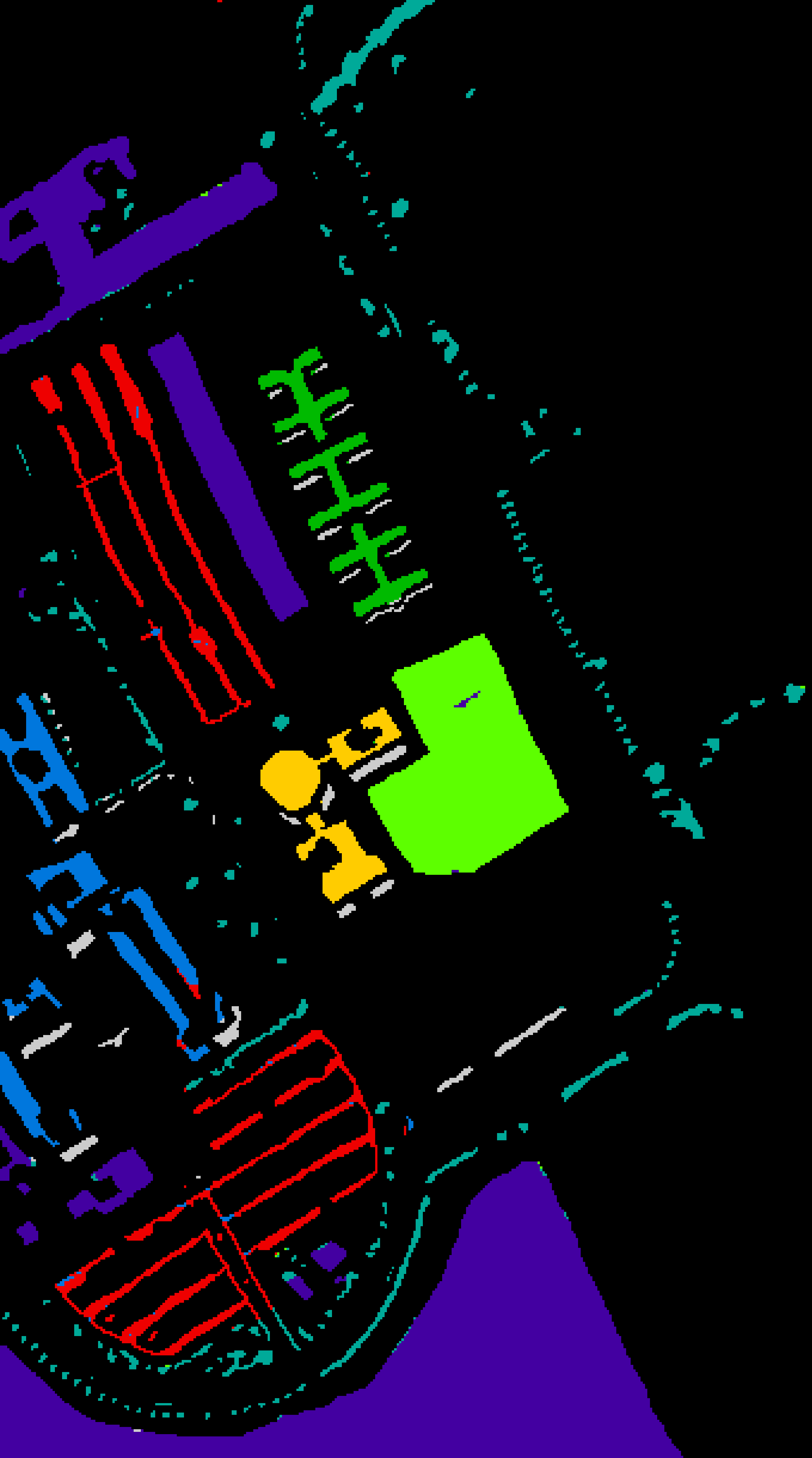}
	\caption*{EF}
    \end{subfigure}
\caption{Classification maps for the \textbf{PU dataset}, highlighting spatial variability and class-specific performance.}
\label{Fig9}
\end{figure}

Table \ref{Tab3} presents the class-wise and aggregate performance of various models on the HC dataset. EF consistently achieves the highest accuracy across multiple categories, with notable improvements in challenging classes such as Watermelon (96.94\%) and Plastic (98.55\%). The OA of 99.28\% and $\kappa$ of 99.16 further confirm its effectiveness. Compared to prior models, EF not only surpasses CNN-based approaches (e.g., HybridSN) but also outperforms state-of-the-art Transformer and Mamba-based architectures, demonstrating its ability to capture complex spectral-spatial dependencies more effectively. Figure \ref{Fig7} provides classification maps for the HC dataset, illustrating EF’s superior spatial consistency and reduced misclassification, particularly in boundary regions. The visual distinction is evident when compared to AGCN, MF, and MHMamba, where spectral confusion and artifacts are more pronounced. The results on the SA and PU datasets, presented in Tables \ref{Tab6} and \ref{Tab5} and Figures \ref{Fig8} and \ref{Fig9}, further reinforce the model’s generalizability. EF attains near-perfect accuracy in multiple categories, outperforming competing methods in complex vegetation and land-cover classes. 

\section{Conclusions}

This paper presents EnergyFormer, a transformer-based framework integrating Multi-Head Energy Attention (MHEA), Fourier Position Embedding (FoPE), and an Enhanced Convolutional Block Attention Module (ECBAM) to achieve state-of-the-art hyperspectral image classification (HSIC) with only 5\% of the available training data. EnergyFormer attains 99.28\% on HC, 98.63\% on SA, and 98.72\% on PU, surpassing CNN, transformer, and Mamba-based models. MHEA mitigates spectral ambiguities, improving accuracy in challenging classes like Watermelon (96.94\%) and Plastic (98.55\%), while FoPE strengthens long-range dependencies and ECBAM enhances feature discrimination. Future work will focus on lightweight deployment and self-supervised learning to tackle data scarcity, ensuring EnergyFormer’s broad applicability in hyperspectral analysis.

\bibliographystyle{IEEEtran}
\bibliography{IEEEabrv,Sam}
\end{document}